\documentclass{article}
\usepackage{tencent_ailab_tech_report}
\usepackage[colorlinks = true,
            linkcolor = blue,
            urlcolor  = blue,
            citecolor = blue,
            anchorcolor = blue]{hyperref}   
\usepackage{microtype}
\usepackage{hyperref}
\usepackage{xurl}
\usepackage{booktabs}
\usepackage{enumitem}
\usepackage{multicol}
\usepackage{CJKutf8}
\usepackage{amsmath}
\usepackage{siunitx}
\usepackage{floatflt}
\usepackage{graphicx}
\usepackage{booktabs}
\usepackage{wrapfig}
\usepackage{authblk}
\usepackage{lipsum}
\usepackage{algorithm}
\usepackage{algorithmicx}
\usepackage{algpseudocode}
\usepackage{microtype}
\usepackage{subfig}
\usepackage{multirow}
\usepackage{booktabs}   
\usepackage{pifont}     
\usepackage{hyperref}
\usepackage{amssymb} 
\algnewcommand{\LeftComment}[1]{\Statex \(\triangleright\) #1}

\usepackage{array}
\usepackage{amsmath}
\usepackage{amssymb}
\usepackage{mathtools}
\usepackage{amsthm}

\usepackage[capitalize,noabbrev]{cleveref}
\usepackage{adjustbox} 

\theoremstyle{plain}

\theoremstyle{definition}

\theoremstyle{remark}

\colmfinalcopy

\usepackage{soul}
\usepackage{fontawesome5}
\usepackage{wrapfig}
\usepackage{caption}
\usepackage{colortbl}
\usepackage{mathabx}
\usepackage{arydshln}

\definecolor{nred}{RGB}{196, 38, 11}
\definecolor{ngreen}{RGB}{18, 141, 21}
\definecolor{nblue}{RGB}{41, 52, 190}
\definecolor{hzw}{RGB}{223, 97, 76}
\definecolor{lt}{RGB}{54, 89, 170}
\definecolor{tblue}{rgb}{0.867, 0.922, 0.969}
\definecolor{zlblue}{RGB}{196, 223, 251}

\newcommand{\ignore}[1]{}
\newcommand{\method}{{DeepMath-103K}}
\newcommand{\methodbf}{\textbf{DeepMath-103K}}

\newcommand{\myhl}[2][yellow!30]{{%
    \sethlcolor{#1}
    \hl{#2}
}}

\def\huggingface{\raisebox{-1.5pt}{\includegraphics[height=1.05em]{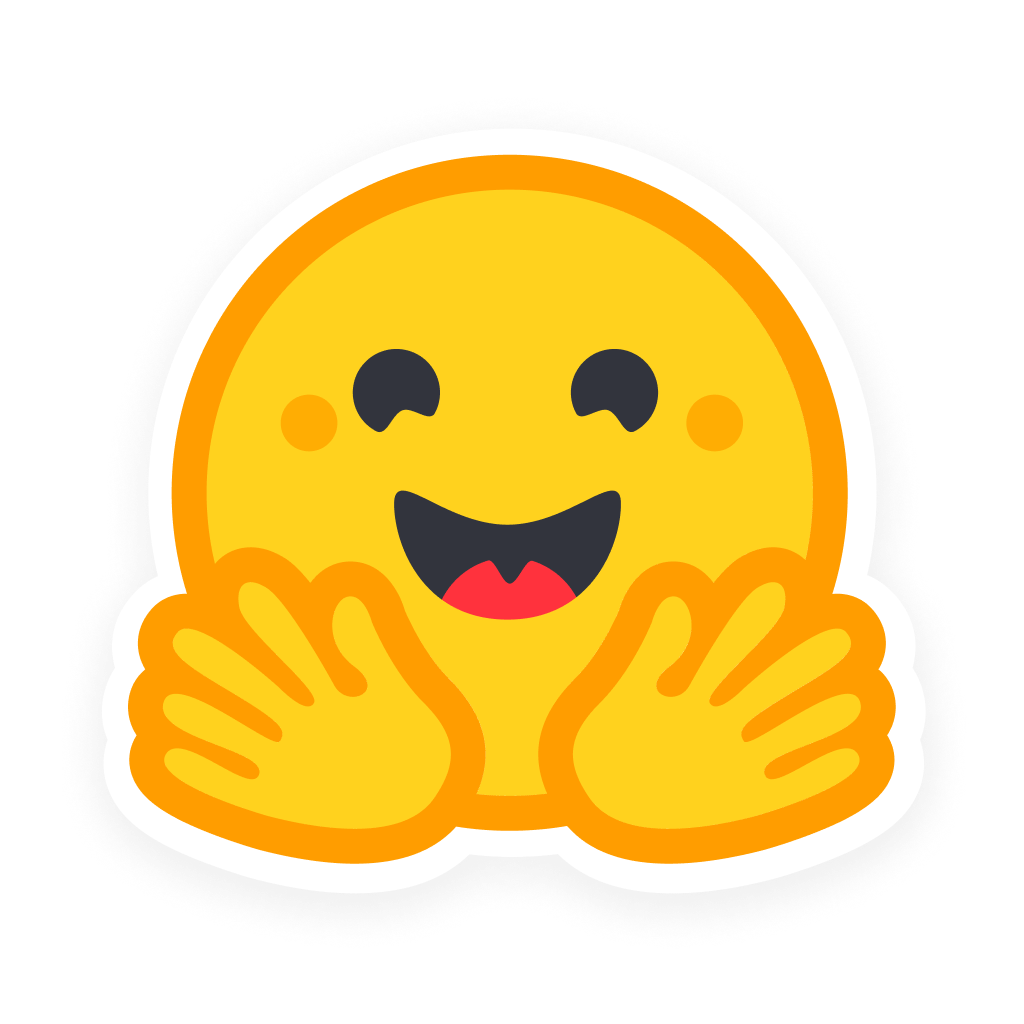}}}
\def\github{\raisebox{-1.5pt}{\includegraphics[height=1.05em]{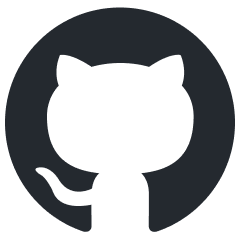}}}

\hypersetup{
    colorlinks=true,   
    linkcolor=purple,  
    citecolor=nblue,   
    filecolor=magenta, 
    urlcolor=nblue     
}

\newlength{\commoncontentheight}

\crefname{section}{\S}{\S\S}
\Crefname{section}{\S}{\S\S}
\crefname{appendix}{appendix}{appendix}

\title{{\em \method:} A Large-Scale, Challenging, Decontaminated, and  Verifiable Mathematical Dataset for Advancing Reasoning}

\author{%
Zhiwei He\thanks{Equal Contribution. The work was done when Zhiwei, Xingyu, and Yue were interning at Tencent.}$^{\phantom{*},1,2}$
~~Tian Liang$^{*,1}$
~~Jiahao Xu$^{*,1}$
~~Qiuzhi Liu$^1$
~~Xingyu Chen$^{1,2}$
~~Yue Wang$^1$\\
\vspace{-9pt}
\bf Linfeng Song$^1$
~~Dian Yu$^1$
~~Zhenwen Liang$^1$
~~Wenxuan Wang$^1$
~~Zhuosheng Zhang$^2$\\
\bf Rui Wang$^{\dag,2}$
~~Zhaopeng Tu\thanks{Correspondence to: Zhaopeng Tu \textless zptu@tencent.com\textgreater~ and Rui Wang \textless wangrui12@sjtu.edu.cn\textgreater.}$^{\phantom{\dag},1}$
~~Haitao Mi$^1$
~~Dong Yu$^1$\\
\vspace{10pt}
$^1$Tencent\ \ \ $^2$Shanghai Jiao Tong University\\
\vspace{10pt}
\hspace{-10pt}\github ~\url{https://github.com/zwhe99/DeepMath}\\
\hspace{-10pt}~~~~~~~~~~~~~~~~~\huggingface ~\url{https://hf.co/datasets/zwhe99/DeepMath-103K}\\
}

\colmfinalcopy 

\begin{document}
\maketitle

\begin{figure}[h!]
\begingroup
\centering
\vspace{-20pt}
\subfloat[Difficulty Levels of different datasets.\label{fig:difficulty}]{
  \begin{minipage}[t]{1\linewidth}
    \adjustimage{width=\linewidth, valign=t}{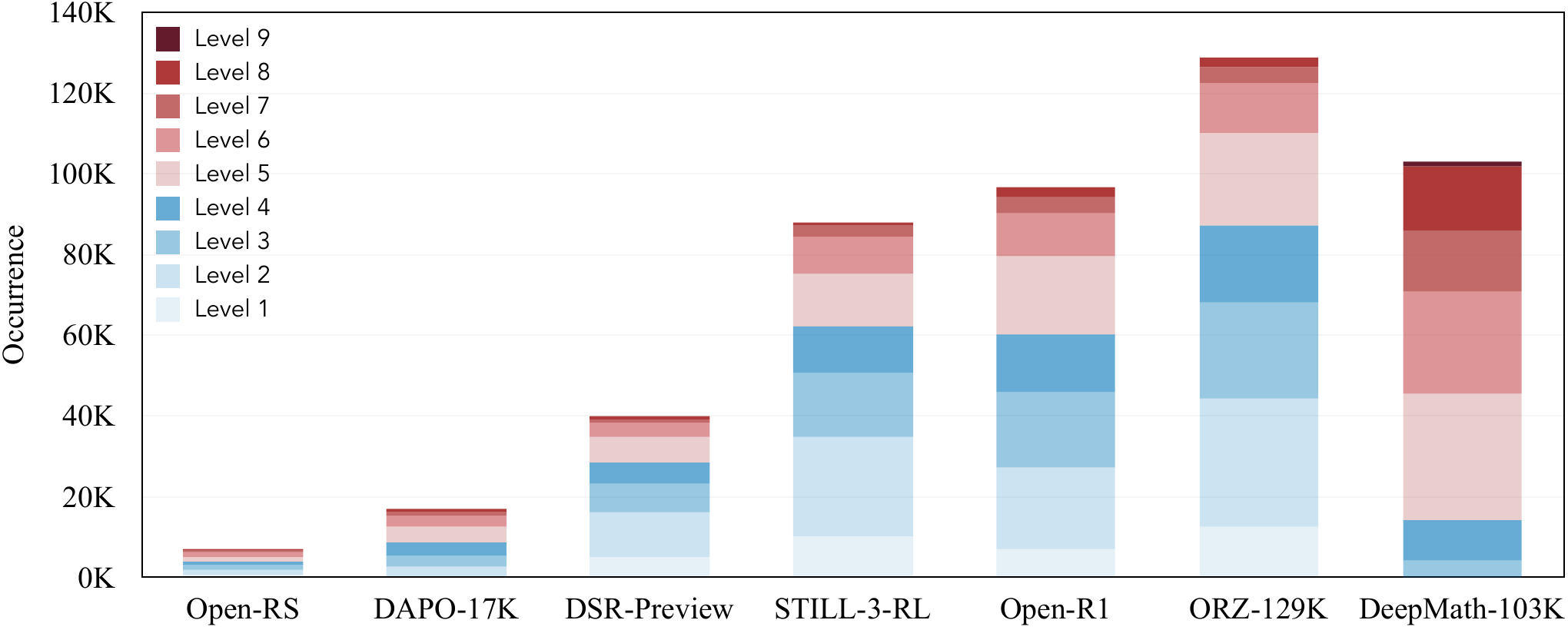}
  \end{minipage}
}
\vspace{10pt}
\subfloat[Pass@1 Accuracies on AIME25.\label{fig:improvement}]{
  \begin{minipage}[t]{1\linewidth}
    \adjustimage{width=\linewidth, valign=t}{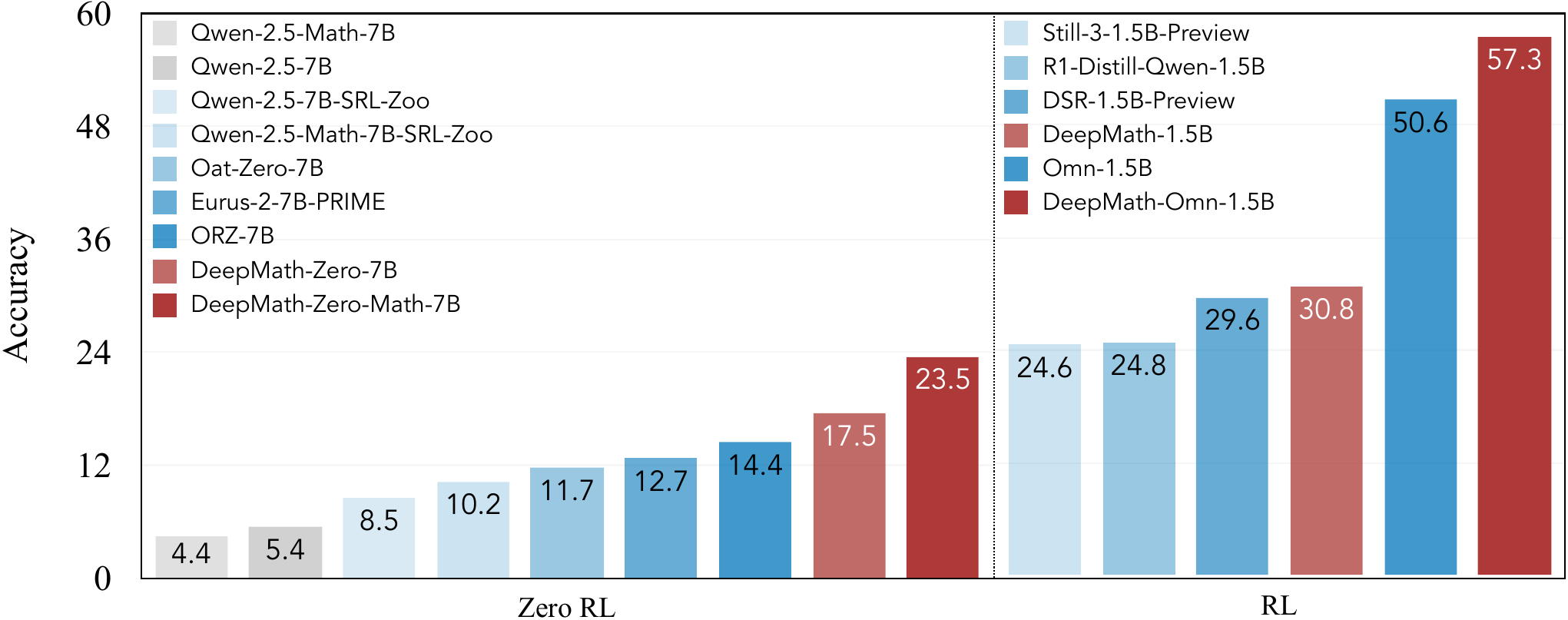}
  \end{minipage}
}
\vspace{5pt}
\caption{(a) \method{} is challenging compared to existing datasets. (b) Results of DeepMath series models under zero RL and RL setting using \method{}.}
\label{fig:abs}
\endgroup
\end{figure}

\begin{abstract}
Reinforcement learning (RL) with large language models shows promise in complex reasoning.
However, its progress is hindered by the lack of large-scale training data that is sufficiently challenging, contamination-free and verifiable.
To this end, we introduce \methodbf{}, a large-scale mathematical dataset designed with high difficulty (primarily levels 5-9), rigorous decontamination against numerous benchmarks, and verifiable answers for rule-based RL reward.
It further includes three distinct R1 solutions adaptable for diverse training paradigms such as supervised fine-tuning (SFT).
Spanning a wide range of mathematical topics, \method{} fosters the development of generalizable and advancing reasoning.
Notably, models trained on \method{} achieve state-of-the-art results on challenging mathematical benchmarks and demonstrate generalization beyond math such as biology, physics and chemistry, underscoring its broad efficacy.
\vspace{30pt}
\end{abstract}

\section{Introduction}
\label{sec:introduction}

Reinforcement learning (RL) with large language models (LLMs) has demonstrated significant potential in complex mathematical reasoning~\citep{guo2025deepseek,OpenReasonerZero2025,zeng2025simplerlzooinvestigatingtamingzero,liu2025understanding}.
Despite this promise, the effective advancement of RL is constrained by existing training data.
While numerous datasets are available, they fall short in several key aspects crucial for training advanced reasoning models: (1) insufficient difficulty (\Cref{fig:difficulty}) to push the boundaries of current models~\citep{dang2025reinforcementlearningreasoningsmall,yu2025dapoopensourcellmreinforcement,deepscaler2025,openr1,OpenReasonerZero2025}, (2) contamination with standard benchmarks (\cref{app:contamination}), (3) a lack of verifiable answers essential for RL with verifiable rewards (RLVR)~\citep{guo2025deepseek, cobbe2021gsm8k, hendrycksmath2021, yumetamath}, or (4) an inadequate combination of these critical aspects at scale.
Furthermore, many of existing datasets represent the recombination and filtration of common sources (such as AIME~\citep{aime}) which contain already well-formatted data, thus lacking a substantial influx of novel and diverse problems from more varied but less structured sources~\citep{dang2025reinforcementlearningreasoningsmall,yu2025dapoopensourcellmreinforcement,deepscaler2025,openr1,OpenReasonerZero2025}.

To bridge this gap, we introduce \methodbf{}, a large-scale mathematical dataset tailored for advancing reasoning via RLVR.
\method{} distinguishes itself through several key features.
\begin{itemize}[leftmargin=15pt, itemsep=3pt, topsep=0pt]
    \item \textbf{Challenging Problems:} \method{} features a high concentration of challenging mathematical problems, with a difficulty distribution skewed towards higher levels ($\geq5$) compared to existing open resources (\Cref{fig:difficulty}).
    \item \textbf{Rigorous Decontamination:} To ensure trustworthy evaluation, \method{} underwent a rigorous decontamination process against a comprehensive suite of benchmarks.
    \item \textbf{Verifiable Answers and Diverse Solutions:} To enable rule-based reward functions in RLVR, every problem in \method{} includes a verifiable final answer that has been validated for easy extraction and verification via rules. Each problem is further enriched with three distinct R1 solutions~\citep{guo2025deepseek}, supporting diverse training paradigms such as SFT.
\end{itemize}

Beyond these core features, \method{} also differentiates itself in its raw data acquisition.
The prevalent trend in existing open datasets often recombines readily available and well-formatted problems from common sources such as AIME~\citep{aime}.
This approach does not create new problems, but re-collect existing ones, which leads to significant overlaps among different datasets.
Recognizing the potential limitations and eventual exhaustion of common resources, \method{} draws its content from more diverse but less structured sources, notably including discussions from Math StackExchange\footnote{\url{https://math.stackexchange.com}}.
The raw content from these sources is informal discourse and lacking a standard format.
After a rigorous curation pipeline that transformed these discussions into a well-structured QA format, \method{} is characterized by its unique problem variety and diversity compared to existing datasets.

Consequently, models trained on \method{} achieve state-of-the-art (SOTA) results (\Cref{fig:improvement}):
\begin{itemize}[leftmargin=15pt, itemsep=3pt, topsep=0pt]
    \item \textbf{Zero RL Training}: Starting from the Qwen-2.5-(Math)-7B~\citep{qwen2.5}, DeepMath-Zero-(Math)-7B shows pass@1 improvements of +12.7 (+23.0) on AIME24 and +12.1 (+19.1) on AIME25, establishing new SOTA performance.
    \item \textbf{RL Training}: Initialized from instruction-tuned models, DeepMath variants also show substantial gains. DeepMath-1.5B, starting from R1-Distill-Qwen-1.5B~\citep{guo2025deepseek}, achieves pass@1 accuracy improvements of +7.9 on AIME24 and +6.0 AIME25. DeepMath-Omn-1.5B, built upon OpenMath-Nemotron-1.5B~\citep{moshkov2025aimo2}, reaches new SOTA pass@1 accuracies of 64.0 on AIME24 and 57.3 on AIME25, surpassing o1-mini (63.6 on AIME24) and low effort o3‑mini (60.0 on AIME24).
    \item \textbf{Generalizable Reasoning beyond Math}: DeepMath series models also generalizes their reasoning abilities to broader domains, achieving best GPQA-Diamond~\citep{gpqa} scores on biology, physics, and chemistry compared to the baselines.
\end{itemize}

These results underscore the value of \method{} as a resource for developing advanced reasoning models with broad applicability.
The remainder of this paper is organized as follows:
\begin{itemize}[leftmargin=15pt]
    \item\Cref{sec:overview} presents an overview of \method{}, including its format, difficulty distribution, and topic covered;
    \item\Cref{sec:curation} details the data curation pipeline to construct \method{}, encompassing source analysis, decontamination, difficulty filtering, and robust answer verification
    \item\Cref{sec:exp} trains, evaluates and analyzes DeepMath series models that trained on \method{}.
\end{itemize}
To foster future research, we have released the \method{} dataset, along with the code and model weights, hoping to enable further exploration of advanced reasoning techniques and the development of robust and generalizable machine intelligence.

\section{Overview of \method}
\label{sec:overview}

\begin{figure}[htpb]
    \centering
    \includegraphics[width=1.0\linewidth]{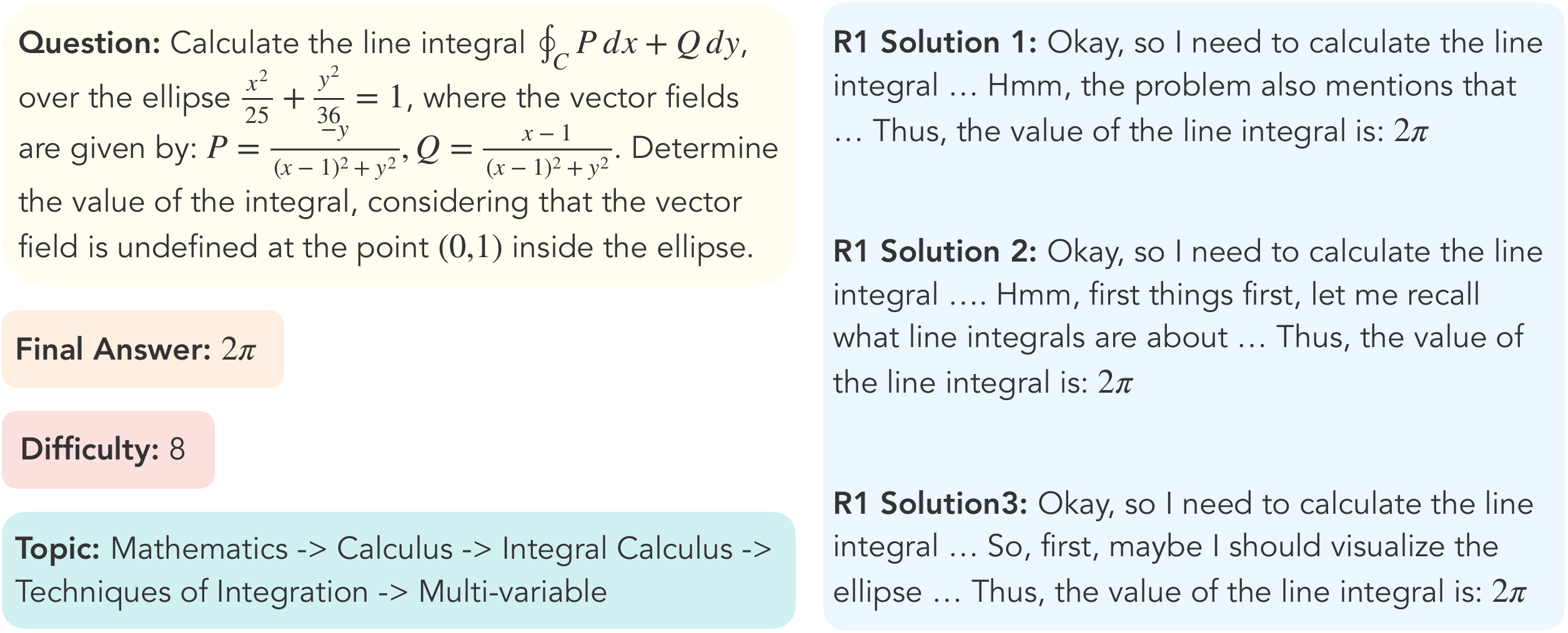}
    \caption{A data sample from \method{}.}
    \label{fig:data-sample}
\end{figure}

Each data sample in \method{} is intentionally structured to be comprehensive, supporting a variety of downstream applications in mathematical reasoning research.
As illustrated in~\Cref{fig:data-sample}, a single sample includes the following components:
\begin{itemize}[leftmargin=15pt]
    \item \textit{Question}: The mathematical problem statement.
    \item \textit{Final Answer}: A verifiable final answer, crucial for rule-based reward functions in RLVR.
    \item \textit{Difficulty}: A numerical difficulty score, which facilitates techniques like difficulty-aware training (e.g., curriculum learning) or adaptive compute allocation based on problem complexity~\citep{wang2025ut, chen2025ot}.
    \item \textit{Topic}: A hierarchical topic classification for the problem, enabling topic-specific analysis.
    \item \textit{R1 Solutions}: Three distinct reasoning paths generated by the DeepSeek-R1 model~\citep{guo2025deepseek}, suitable for diverse training paradigms such as SFT.
\end{itemize}

\method{} possesses several key characteristics that make it particularly suitable for advancing mathematical reasoning research:

\paragraph{Higher Difficulty}
\method{} includes mathematical problems spanning difficulty levels 3 through 9.
The core of the dataset consists of 95K challenging problems (levels 5-9) specifically curated for this research.
To ensure broader difficulty coverage, this is augmented with an additional 8K problems (levels 3-5) sourced from SimpleRL~\citep{zeng2025simplerl}.
For comparison, we analyzed and labeled the difficulty levels of several existing datasets commonly used for RLVR training in math domain: Open-RS~\citep{dang2025reinforcementlearningreasoningsmall}, DAPO-17K~\citep{yu2025dapoopensourcellmreinforcement}, DSR-Preview~\citep{deepscaler2025}, SITLL-3-RL~\citep{sitll3_1}, ORZ-129K~\citep{OpenReasonerZero2025}, and Open-R1~\citep{openr1}.
\Cref{fig:difficulty} illustrates the difficulty distributions across these datasets.
As depicted, \method{} exhibits a significantly more challenging problem distribution, containing a substantially higher proportion of problems at difficulty levels 5 and above compared to the other benchmark datasets.
This focus on higher difficulty is intended to push the reasoning limits of current models.

\paragraph{Rigorous Data Decontamination}
\begin{figure}[b]
    \centering
    \includegraphics[width=1\linewidth]{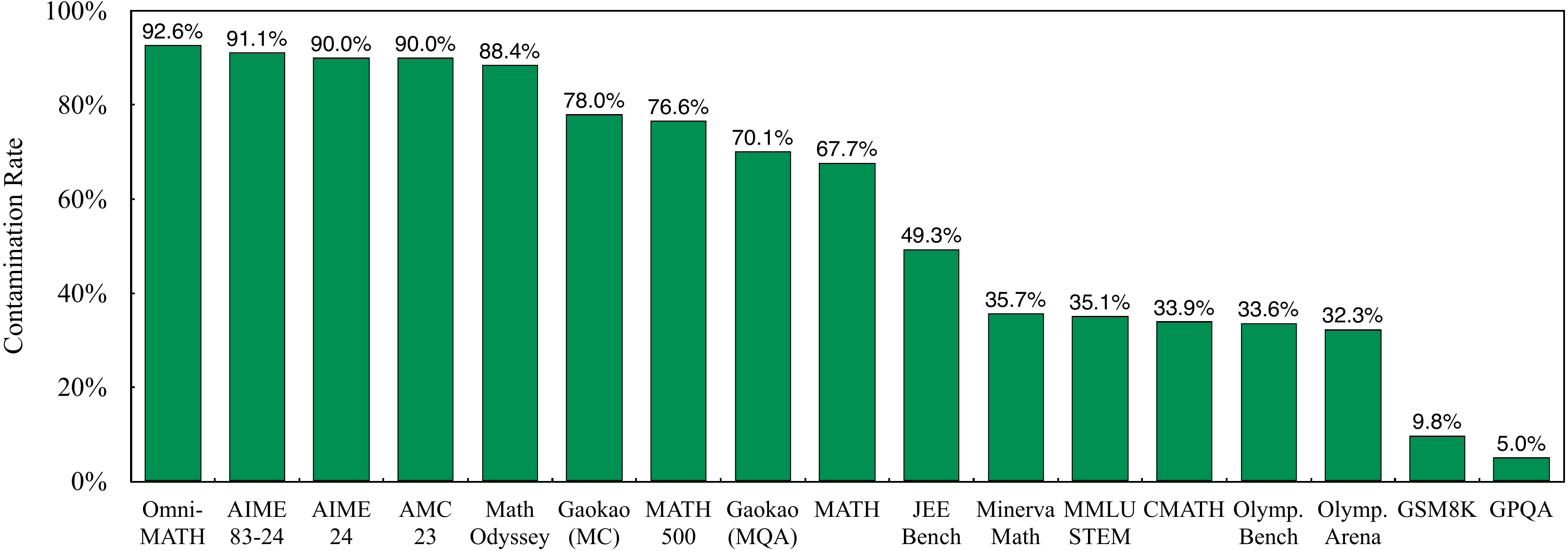}
    \caption{Contamination rates of common mathematical and STEM benchmarks detected in the raw data sources before decontamination.}
    \label{fig:contamination-dist}
\end{figure}
\method{} was constructed exclusively using the training splits of existing open resources, with careful avoidance of any known test set materials.
However, our preliminary analysis revealed that these source data exhibits alarmingly high levels of contamination with problems from commonly used evaluation benchmarks.
As illustrated in~\Cref{fig:contamination-dist}, the contamination rates (defined as the percentage of benchmark test samples found within our raw data pool) are notably high: reaching 90\% for AIME24 and AMC23, 76.6\% for MATH500, 35.7\% for Minerva Math, and 33.6\% for OlympiadBench.
Recognizing that these benchmarks are frequently employed for model evaluation, \method{} underwent a rigorous decontamination procedure.
This process systematically identified and removed problems that overlap with these standard evaluation sets, ensuring the integrity and reliability of future benchmark results obtained using models trained on \method{}.

\paragraph{Broad Topical Diversity}
\begin{wrapfigure}{r}{0.50\linewidth}
\vspace{2.5pt}
\centering
    \includegraphics[width=\linewidth]{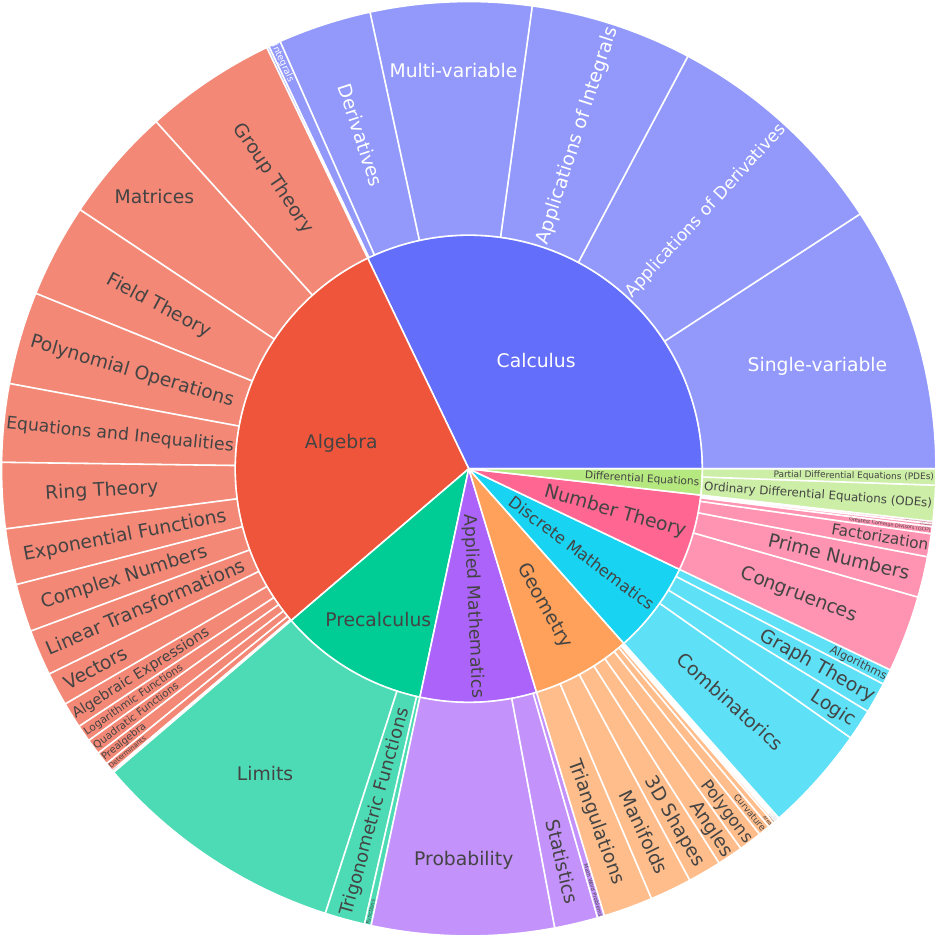}
    \caption{Hierarchical breakdown of covered mathematical topics in \method.}
    \label{fig:domain}
    \vspace{-11pt}
\end{wrapfigure}
Complementing its high difficulty and data integrity, a key characteristic of \method{} is its extensive topical diversity spanning the mathematical landscape.
We categorized each problem using a hierarchical topic structure, following the methodology from~\citet{gao2024omnimathuniversalolympiadlevel}.
As illustrated in~\Cref{fig:domain}, this classification reveals that \method{} draws problems from a multitude of core mathematical areas.
Its scope ranges from fundamental topics such as Prealgebra and Plane Geometry to sophisticated domains like Abstract Algebra (including Group Theory and Field Theory) and advanced Calculus (covering Differential Equations and Applications of Integrals, among others).
This broad and deep topical foundation ensures that models trained on \method{} are exposed to a rich variety of mathematical concepts and problem-solving paradigms, thereby fostering the development of more robust and widely generalizable reasoning skills.

\begin{figure}[htpb]
    \centering
    \includegraphics[width=\linewidth]{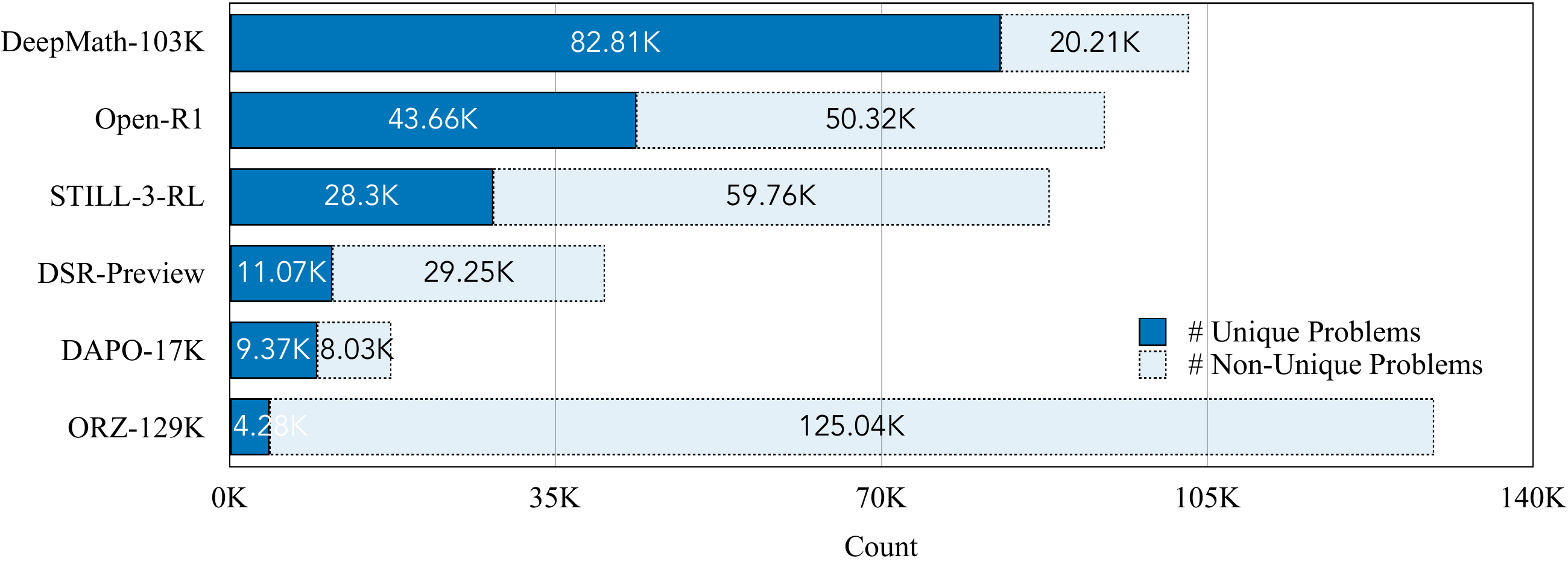}
    \caption{Unique and non-unique problem counts in \method{} compared to other datasets.}
    \label{fig:data-novelty-bar}
\end{figure}

\begin{figure}[t]
    \centering
    \subfloat[\method{}\label{fig:tsne-deepmath103k}]{
      \begin{minipage}[t]{0.276\linewidth}
        \adjustimage{width=\linewidth, valign=t}{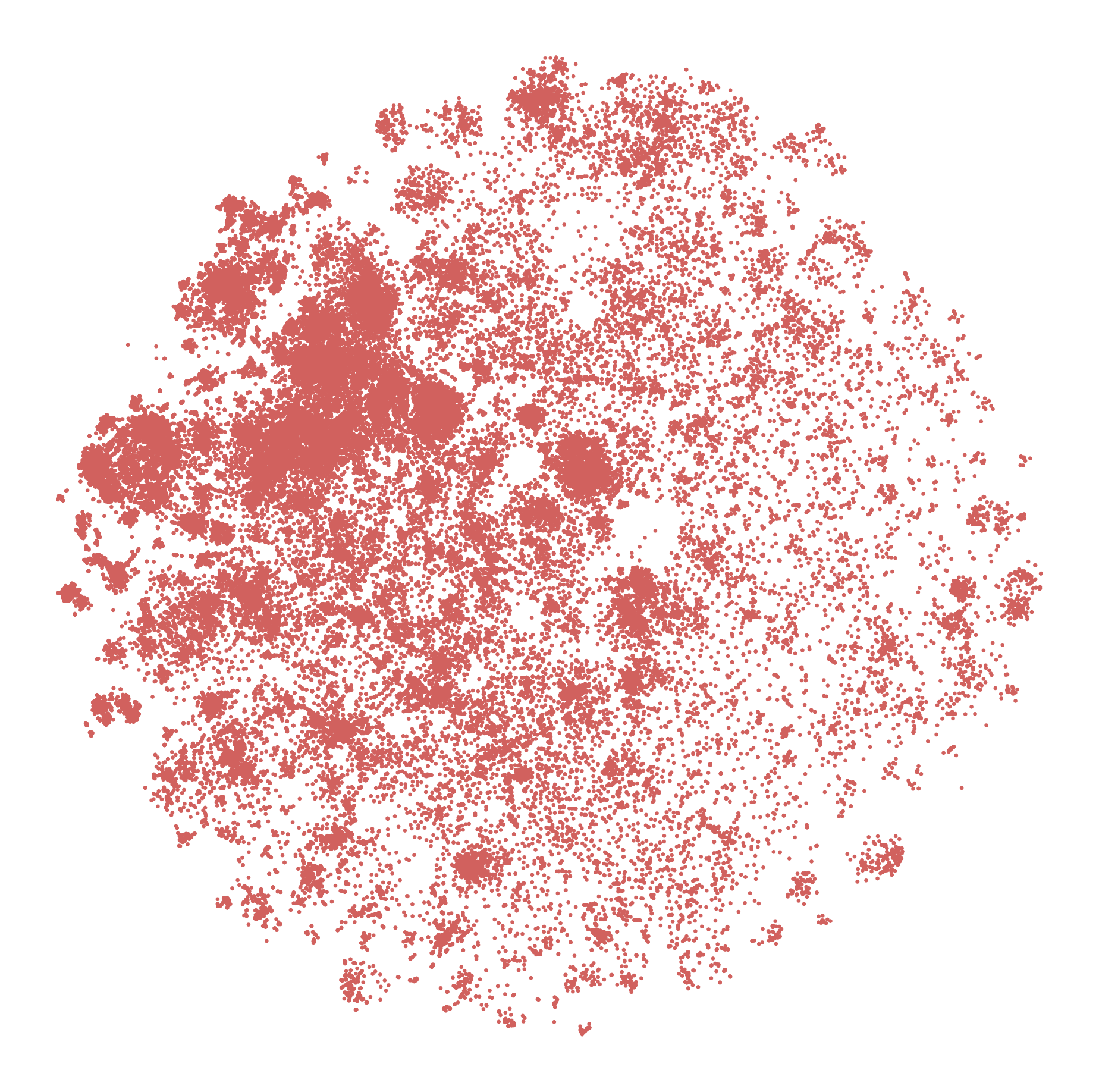}
      \end{minipage}
    }
    \subfloat[ORZ-129K\label{fig:tsne-orz129k}]{
      \begin{minipage}[t]{0.276\linewidth}
        \adjustimage{width=\linewidth, valign=t}{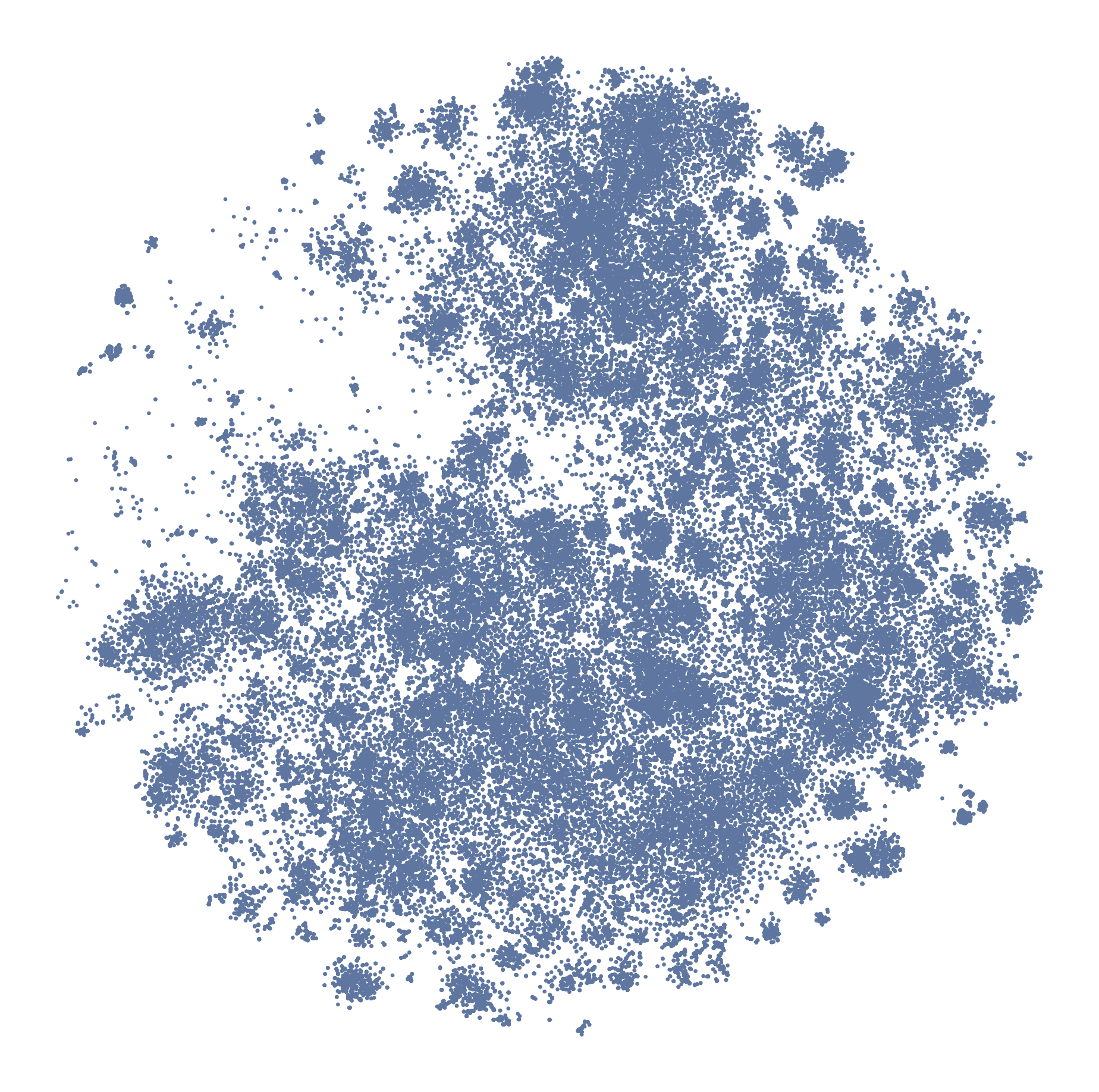}
      \end{minipage}
    }
    \subfloat[Open-R1\label{fig:tsne-openr1}]{
      \begin{minipage}[t]{0.276\linewidth}
        \adjustimage{width=\linewidth, valign=t}{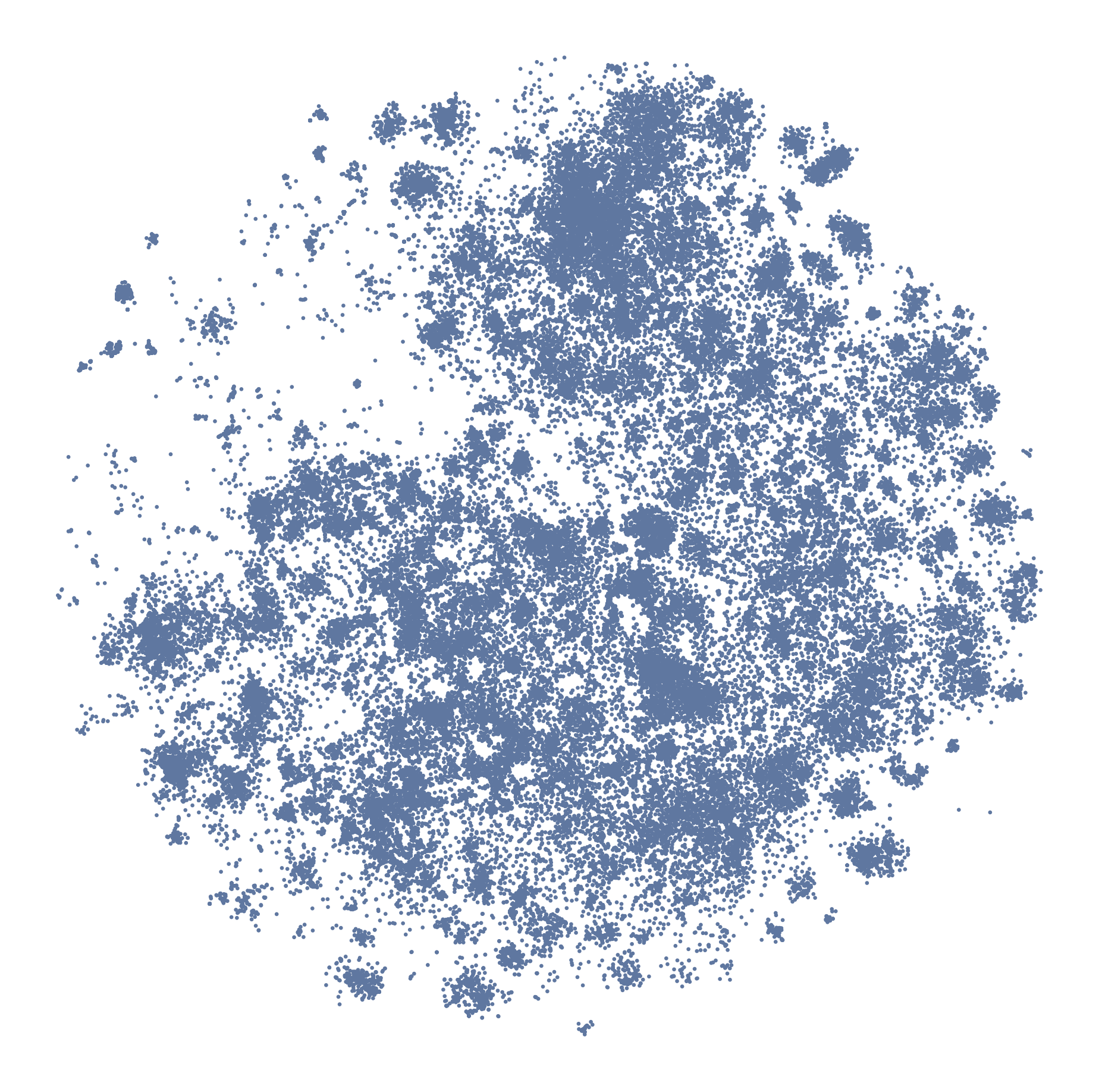}
      \end{minipage}
    }
    \vfill
    \subfloat[STILL-3-RL]{
      \begin{minipage}[t]{0.276\linewidth}
        \adjustimage{width=\linewidth, valign=t}{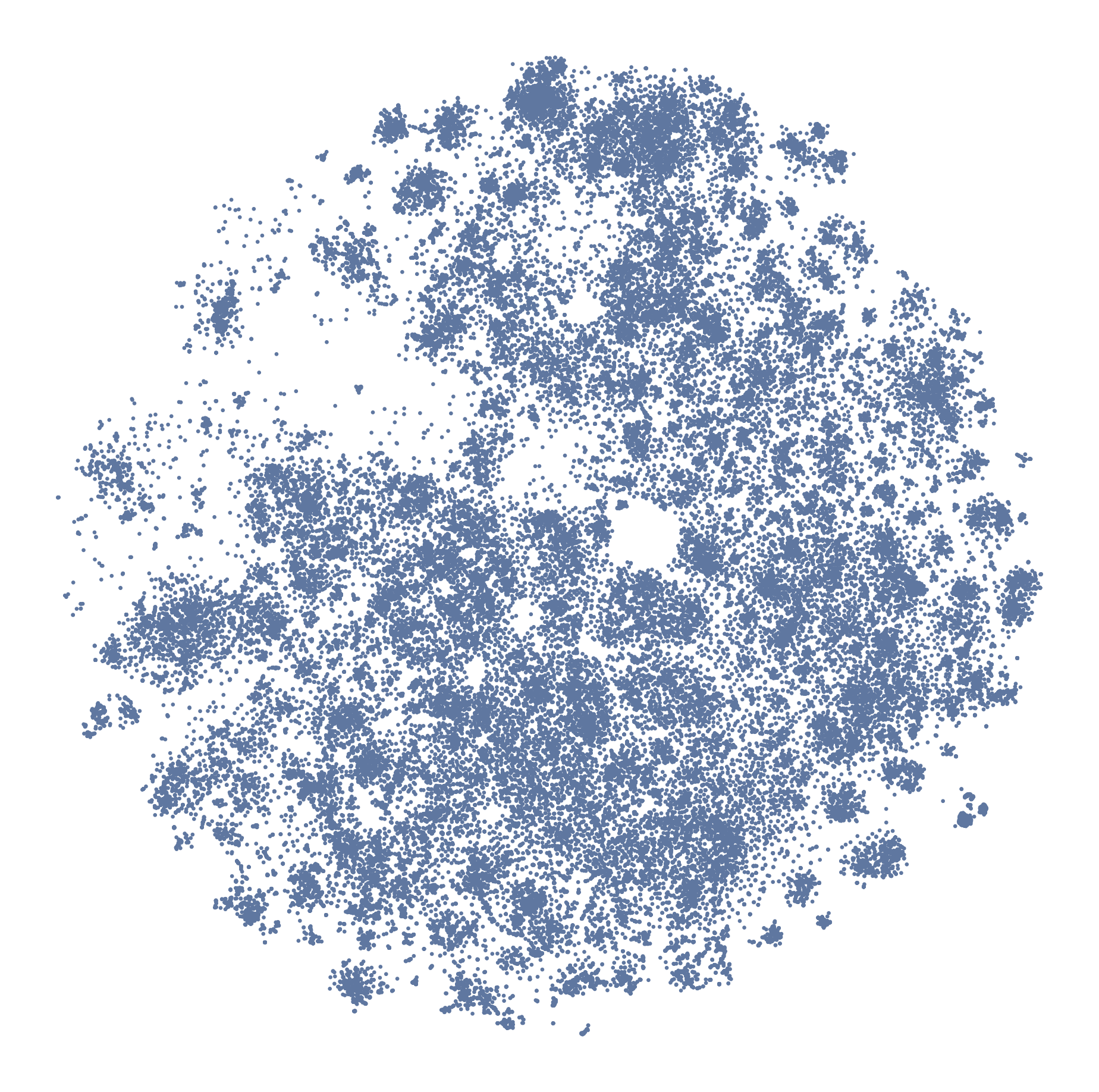}
      \end{minipage}
    }
    \subfloat[DSR-Preview]{
      \begin{minipage}[t]{0.276\linewidth}
        \adjustimage{width=\linewidth, valign=t}{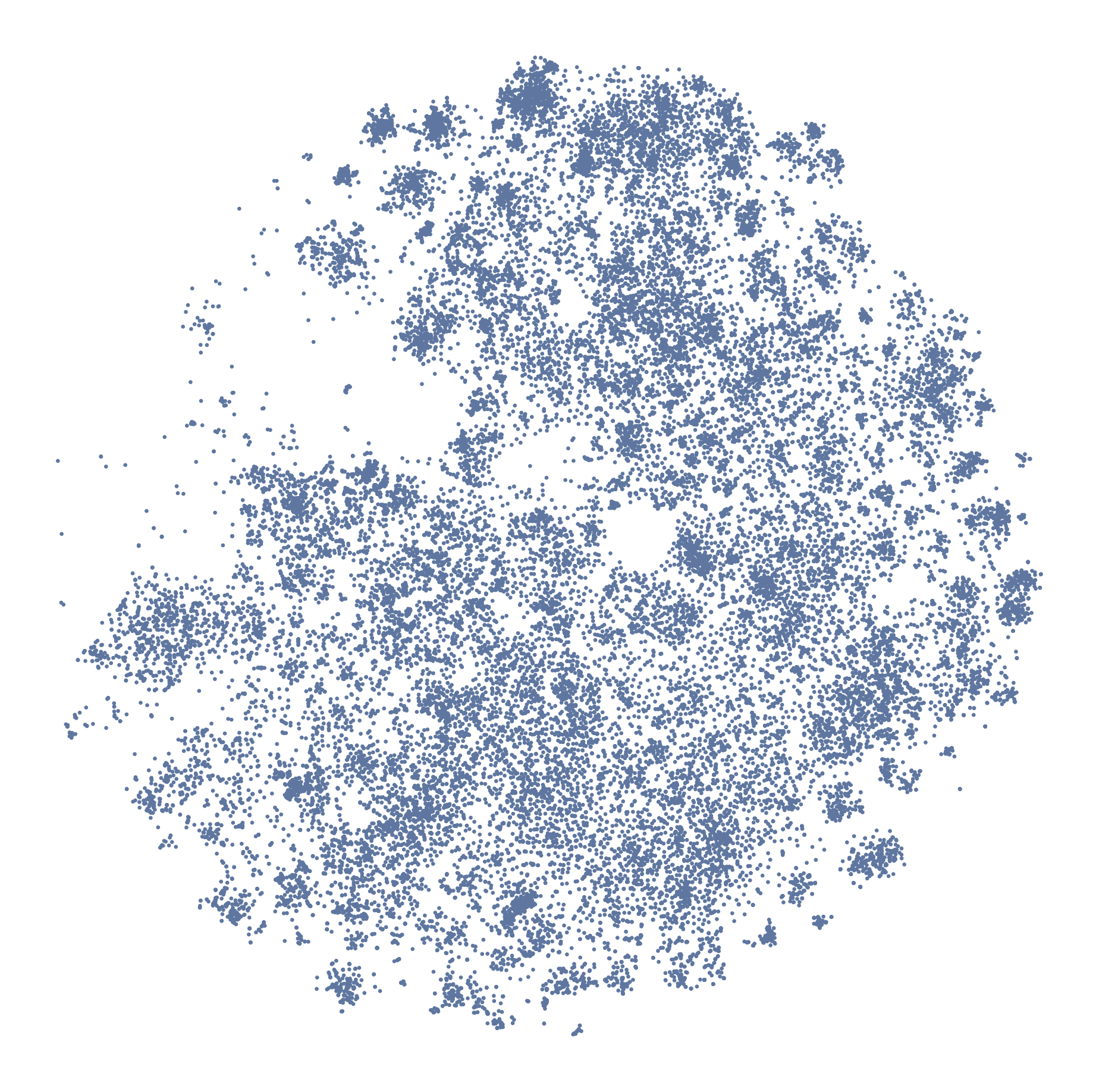}
      \end{minipage}
    }
    \subfloat[DAPO-17K]{
      \begin{minipage}[t]{0.276\linewidth}
        \adjustimage{width=\linewidth, valign=t}{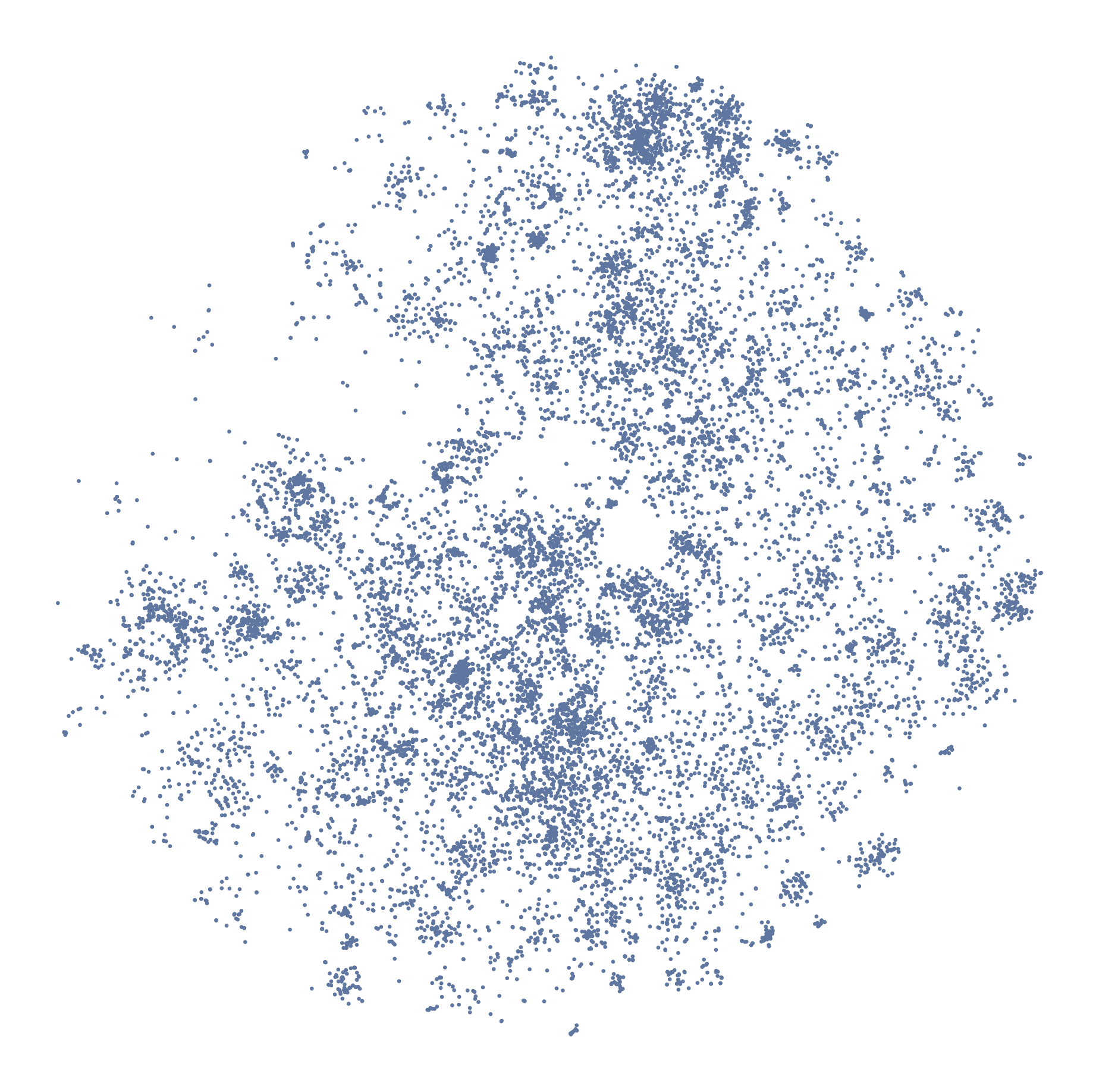}
      \end{minipage}
    }

    \caption{Embedding distributions of different datasets after t-SNE.}
    \label{fig:tsne}
\end{figure}

\paragraph{Data Novelty and Uniqueness}
As mentioned in~\Cref{sec:introduction}, \method{} sources mostly from math forum, rather than common resources frequently adopted by other datasets.
To evaluate the data novelty and uniqueness of~\method{}, we performed the following analysis for all the datasets:
\begin{enumerate}[leftmargin=15pt, itemsep=3pt, topsep=0pt]
    \item We first embedded all the samples using paraphrase-multilingual-MiniLM-L12-v2.
    \item Samples with an embedding similarity greater than 0.98 were considered as the same samples.
\end{enumerate}
Viewing each dataset as a set of embeddings, \Cref{fig:data-novelty-bar} presents the number of unique elements in each set and the corresponding set sizes.
\method{} contains 82.81K problems that are not found in others.
This stark contrast highlights the data novelty and uniqueness of \method.
We also plot their embedding distribution after t-SNE in~\Cref{fig:tsne}.
ORZ-129K~\citep{OpenReasonerZero2025}, Open-R1~\citep{openr1}, SITLL-3-RL~\citep{sitll3_1}, DSR-Preview~\citep{deepscaler2025}, and DAPO-17K~\citep{yu2025dapoopensourcellmreinforcement}, though curated independently, show very similar embedding distribution, while \method{} exhibits a distinctly different pattern.
This observation supports our claim that existing datasets overlap with each other because of using common data sources and further demonstrate the data novelty and uniqueness of \method{}.

\section{Construction of \method}
\label{sec:curation}

\begin{figure}[htpb]
    \vspace{-13pt}
    \centering
    \includegraphics[width=\linewidth]{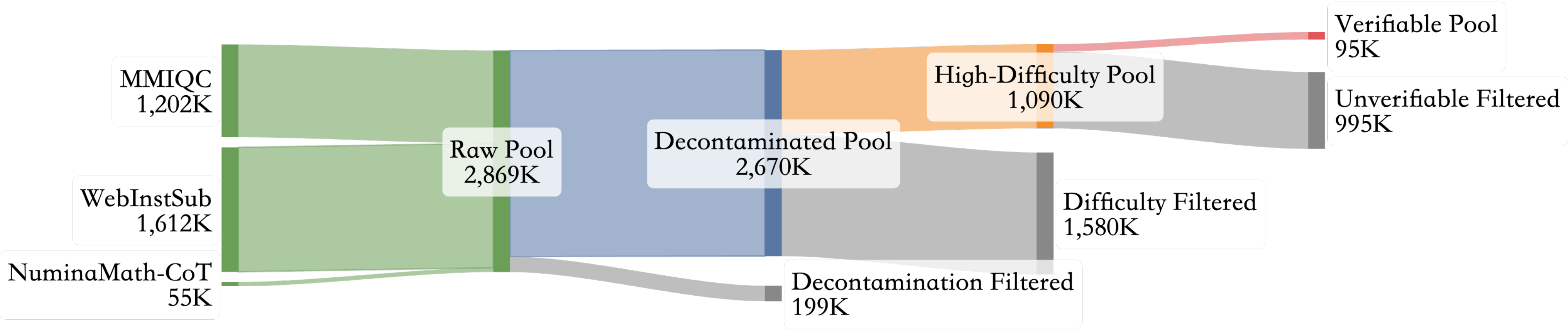}
    \caption{The data curation pipeline for \method{}. Starting with an initial pool of 2,869K raw questions, successive stages of data decontamination, difficulty filtering (retaining levels $\geq$5), and answer verifiability filtering yield 95K problems. These are then combined with 8K problems from SimpleRL~\citep{zeng2025simplerl} to form the final \method{} dataset.}
    \label{fig:data-curation-pipeline}
\end{figure}
This section details the meticulous data curation process used to construct \method{}, illustrated in Figure~\ref{fig:data-curation-pipeline}. The process comprises four primary stages:
\begin{enumerate}[itemsep=0pt,topsep=0pt]
    \item \textbf{Source Analysis and Collection:} Identifying and collecting mathematically challenging problems by analyzing the difficulty distributions of existing open data sources.
    \item \textbf{Data Decontamination:} Rigorously decontaminating the collected data to remove potential overlaps with standard evaluation benchmarks, ensuring evaluation integrity.
    \item \textbf{Difficulty Filtering:} Filtering the decontaminated problems based on difficulty, retaining only those assessed at level 5 or higher to focus on challenging content.
    \item \textbf{Answer Verification:} Ensuring each curated problem possesses a verifiable final answer, consistently validated across multiple solution paths generated by DeepSeek-R1.
\end{enumerate}
Overall, this curation pipeline ensures that \method{} is largely free from benchmark contamination and concentrates on challenging mathematical problems suitable for advanced reasoning model training.
The entire procedure involved significant computational resources, requiring an expenditure of {\bf 138,000 US dollars in GPT-4o API fees} and a total of {\bf 127,000 H20 GPU hours}.

\begin{wrapfigure}{r}{0.39\linewidth}
\centering
    \vspace{-13pt}
    \includegraphics[width=\linewidth]{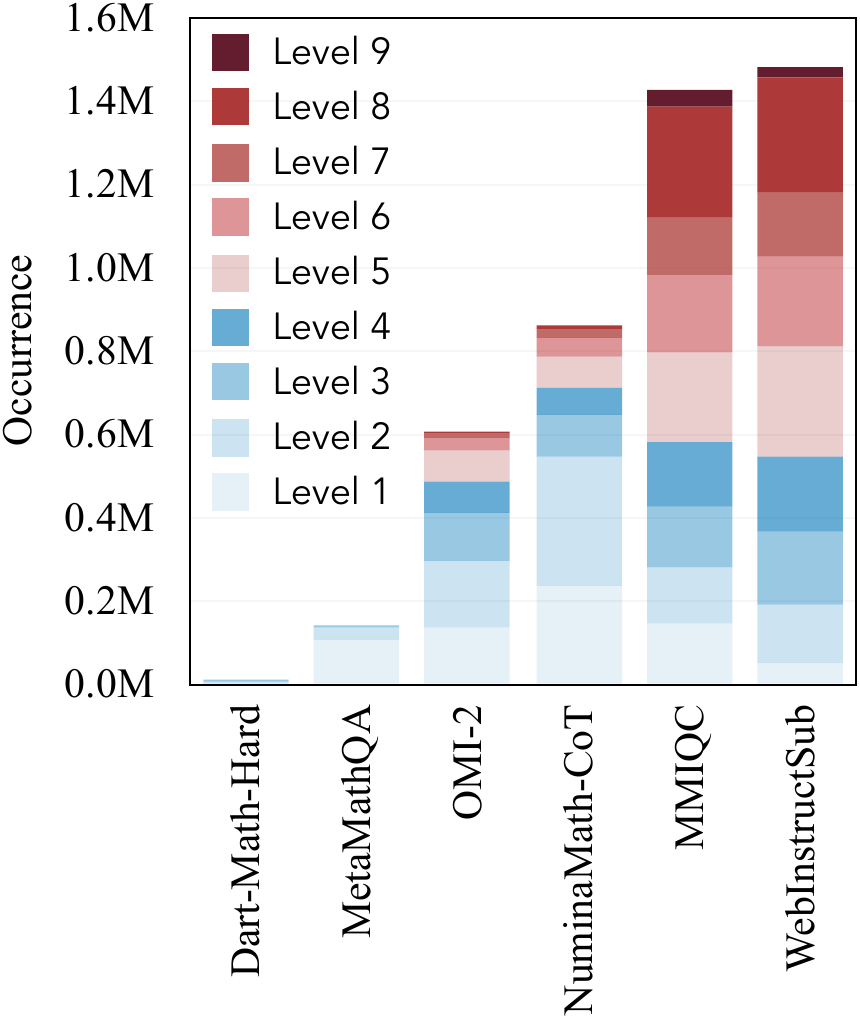}
    \vspace{-21pt}
    \caption{Difficulty distributions of various open mathematical datasets considered as potential sources.}
    \label{fig:raw_data_difficulty}
    \vspace{-10pt}
\end{wrapfigure}

\paragraph{Stage 1: Source Analysis and Collection.}
To identify data sources rich in challenging problems, we began by analyzing the landscape of existing open mathematical reasoning datasets designed for SFT.
These datasets utilize diverse collection methods.
For instance, datasets such as MetaMathQA~\citep{yumetamath}, dart-math-hard~\citep{tong2024dartmath}, and OpenMathInstruct-2~\citep{toshniwal2024openmath2} primarily focus on augmenting problems and solutions derived from established datasets like GSM8K~\citep{cobbe2021gsm8k} and MATH~\citep{hendrycksmath2021}.
In contrast, datasets like NuminaMath-CoT~\citep{numina_math_datasets}, 
MMIQC~\citep{liu2024augmentingmathwordproblems}, and WebInstructSub~\citep{yue2024mammoth2} source content more broadly from the web, gathering materials such as exercises and discussions from online platforms (e.g., Math Stack Exchange).
We follow~\citet{gao2024omnimathuniversalolympiadlevel} to estimate the difficulty distributions of these potential source datasets, as shown in~\Cref{fig:raw_data_difficulty}, which reveals distinct patterns: datasets derived from GSM8K and MATH (MetaMathQA, dart-math-hard, OpenMathInstruct-2), along with NuminaMath-CoT, exhibited distributions heavily skewed towards lower difficulty levels (levels 1-5).
Conversely, datasets sourced more broadly from web content, specifically MMIQC and WebInstructSub, displayed significantly flatter distributions with a larger proportion of problems in the mid-to-high difficulty range (levels 5-9).
Based on this finding, we selected Math StackExchange subsets from
MMIQC and WebInstructSub as our primary data sources due to their higher concentration of challenging problems.
We also included NuminaMath-CoT to enhance the topical diversity of the initial collection. After applying basic filtering, this selection process yielded a raw pool of 2,869K questions.

\paragraph{Stage 2: Data Decontamination.}
As indicated by the high contamination rates observed in common benchmarks (\Cref{fig:contamination-dist}), a rigorous data decontamination process was crucial for ensuring the integrity of \method{}.
We performed decontamination against a comprehensive suite of mathematics and STEM benchmarks, including MATH~\citep{hendrycksmath2021}, AIME~\citep{aime}, AMC~\citep{amc}, Minerva Math~\citep{minerva}, OlympiadBench~\citep{he2024olympiadbench}, Omni-MATH~\citep{gao2024omnimathuniversalolympiadlevel}, MathOdyssey~\citep{fang2024mathodyssey}, GAOKAO~\citep{zhong2023agieval}, JEEBench~\citep{arora-etal-2023-llms}, MMLU-STEM~\citep{hendryckstest2021}, CMATH~\citep{wei2023cmath}, OlympicArena~\citep{huang2024olympicarena}, GSM8K~\citep{cobbe2021gsm8k}, and GPQA~\citep{gpqa}.
We adopted the decontamination method proposed by~\citet{toshniwal2024openmath2}:
\begin{enumerate}[leftmargin=15pt, itemsep=3pt, topsep=0pt]
    \item For each candidate question in our raw dataset, we employed embedding similarity search (using paraphrase-multilingual-MiniLM-L12-v2~\citep{reimers-2019-sentence-bert}) to identify the top-$k$ ($k=5$) most similar examples from the aggregated test sets of all targeted benchmarks.
    \item Each candidate question was then compared against its top-$k$ retrieved benchmark examples using an LLM-Judge (Llama-3.3-70B-Instruct~\citep{grattafiori2024llama}) to determine if they constituted identical questions or paraphrases. If any of these comparisons indicated a potential paraphrase or duplicate, the candidate question was discarded.
\end{enumerate}
\Cref{tab:decontamination-examples} illustrates the effectiveness of semantic decontamination compared to simple lexical matching.
This approach aims to identify not only exact duplicates but also near-duplicates and paraphrased questions that might otherwise overlap with evaluation sets.
\begin{table}[t]
    \centering
    \caption{Examples of contamination detected between the raw data pool and benchmarks using semantic comparison. Colors highlight conceptual or textual similarities.}
    \resizebox{\linewidth}{!}{
    \begin{tabular}{p{0.10\linewidth} p{0.45\linewidth} p{0.45\linewidth}}
    \toprule
    \textbf{Benchmark} & \textbf{Raw Question} & \textbf{Benchmark Question} \\
    \midrule
    AIME24 & \myhl[yellow!20]{How many routes} are there through from \myhl[red!25]{top left corner to bottom right} in a \myhl[zlblue]{20x20 grid}? I'm trying to solve this computer programming problem on Project Euler. I've seen a solution using nCr, where n = 40 and r = 20. Could someone explain to me how this work, please? & Consider the paths of length 16 that follow the lines from the \myhl[red!25]{lower left corner to the upper right corner} on an \myhl[zlblue]{8x8 grid}. \myhl[yellow!20]{Find the number of such paths} that change direction exactly four times, as in the examples shown below.\\
    \midrule
    AMC23 & Using only \myhl[yellow!20]{3 paise, 5 paise, and 9 paise coins}, what is the \myhl[red!25]{largest amount} that \myhl[zlblue]{cannot be paid in exact change}?  & In the state of Coinland, coins have values \myhl[yellow!20]{6,10, and 15 cents}. Suppose \myhl[red!25]{x is the value in cents of the most expensive item} in Coinland that \myhl[zlblue]{cannot be purchased using these coins with exact change}. What is the sum of the digits of x?\\
    \bottomrule
    \end{tabular}}
    \label{tab:decontamination-examples}
\end{table}

\paragraph{Stage 3: Difficulty Filtering.}
\begin{table}[htpb]
    \vspace{-10pt}
    \centering
    \caption{Examples of geometry problems retained by the difficulty filtering process (level $\geq$ 5).}
    \resizebox{\linewidth}{!}{
    \begin{tabular}{p{0.08\linewidth} p{0.92\linewidth}}
    \toprule
    \textbf{Difficulty} & \textbf{Problem} \\
    \midrule
    5 & Four random points are placed in the plane, with each point's horizontal and vertical coordinates uniformly distributed on the interval $(0,1)$. What is the expected largest size of a subset of these points that can form the vertices of a convex polygon? \\
    \midrule
    6 & A square has one side lying on the line $y = 2x - 17$ and two other vertices on the parabola $y = x^2$. Determine the minimum possible area of the square."\\
    \midrule
    7 & Determine the sequence $s(k,n)$, which represents the number of sides of the intersection of a unit-radius regular polygon $P_k$ with $k$ sides and a rotating unit-radius regular polygon $P_n$ with $n \ge k$ sides, as the angle of rotation $\theta$ varies from $0$ to $2\pi$. Provide the sequence $s(k,n)$ for all $n \ge k$. \\
    \midrule
    8 & Consider a convex n-gon $A_1 A_2 \cdots A_n$ inscribed in a unit circle. Determine the maximum value of the sum of the squares of all its sides and diagonals \\
    \midrule
    9 & Determine the maximal cardinality of a collection $ \mathcal{C} $ of projective planes on $ \omega $ such that no two distinct members of $ \mathcal{C} $ are isomorphic. A set $ L \subseteq \mathcal{P}(X) $ is a projective plane on $ X \neq \emptyset $ if: 1. For any distinct $ x, y \in X $, there is a unique $ l \in L $ such that $ x, y \in l $. 2. For any distinct $ l, m \in L $, $ |l \cap m| = 1 $. 3. There exist four distinct elements of $ X $ such that no member of $ L $ contains more than two of these four elements. Two projective planes $ L $ and $ M $ on $ X $ are isomorphic if there is a bijection $ \varphi: X \to X $ such that $ l \in L $ if and only if $ \varphi(l) \in M $.\\
    \bottomrule
    \end{tabular}}
    \label{tab:difficuly-examples}
\end{table}
\citet{zeng2025simplerlzooinvestigatingtamingzero} highlights the importance of aligning RL training data difficulty with the target model's reasoning capabilities, noting that powerful models benefit significantly from exposure to challenging problems.
Building on this insight, our curation process for \method{} focuses on selecting problems that represent a significant reasoning challenge.
To quantify difficulty, we adopted the approach detailed in Omni-MATH~\citep{gao2024omnimathuniversalolympiadlevel}.
We assigned a difficulty level to each decontaminated problem by prompting GPT-4o based on the annotation guidelines provided by the AoPS\footnote{\url{https://artofproblemsolving.com/wiki/index.php/AoPS\_Wiki:Competition\_ratings}}.
To ensure a robust estimate, we queried GPT-4o six times for each problem and averaged the resulting ratings to determine its final difficulty level.
Subsequently, we applied a strict filtering criterion, retaining only those problems with an estimated difficulty level of 5 or higher.
\Cref{tab:difficuly-examples} showcases examples of geometry problems that passed this filtering stage, illustrating how increasing difficulty levels often correlate with greater conceptual depth and reasoning complexity.

\paragraph{Stage 4: Answer Verification.}
The availability of verifiable final answers is crucial for enabling rule-based reward in RLVR, which helps mitigate reward hacking and has been instrumental in training successful reasoning models like DeepSeek-R1~\citep{guo2025deepseek}. However, reliably constructing such answers presents two primary challenges:
\begin{enumerate}[leftmargin=15pt, itemsep=3pt, topsep=0pt]
    \item Some open-ended questions inherently lack a easily verifiable final answer.
    \item Certain answers are excessively complex (e.g., lengthy expressions or intricate notation), making them challenging or even infeasible for automated rule-based verification.
\end{enumerate}
To address these issues, we implemented a rigorous two-stage verification process:
\begin{enumerate}[leftmargin=15pt, itemsep=3pt, topsep=0pt]
    \item \textbf{Question Filtering and Formatting:} We used GPT-4o to process the raw questions. Problem types inherently unsuitable for verification were discarded. Questions phrased conversationally were rewritten into a standardized format seeking a single, specific numerical or symbolic answer.
    \item \textbf{Answer Verification via Consistency Check:} For questions successfully passing the above step, we generated three distinct solution paths using DeepSeek-R1. A rule-based verifier then extracted the final answer from each of these generated solutions, as well as from the original source solution (when available). We enforced strict consistency: only problems where all extracted final answers were identical were retained in the final dataset.
\end{enumerate}
This combined approach of question standardization and multi-solution answer consistency checking ensures that every problem included in \method{} possesses a final answer that is robustly verifiable using automated rules.
\section{DeepMath Series Models}
\label{sec:exp}
This section presents a comprehensive evaluation of the mathematical and general reasoning capabilities of our DeepMath series of models, which were trained on \method{}.

\subsection{Setup}
\paragraph{Training Paradigms} We employed two distinct RL training paradigms:
\begin{itemize}[leftmargin=10pt, itemsep=3pt, topsep=0pt]
    \item \textit{Zero RL:} This paradigm involves training LLMs from their base (non-instruction-tuned) version using RL~\citep{guo2025deepseek}. We used group relative policy optimization (GRPO)~\citep{shao2024deepseekmathpushinglimitsmathematical} with fixes from~\citet{yu2025dapoopensourcellmreinforcement}, and trained Qwen-2.5-(Math)-7B with a rule-based reward (+1 for correct final answer, -1 otherwise). Detailed settings are available in~\Cref{app:training-detail}, and SFT results are presented in~\Cref{sec:sft-results}.
    \item \textit{RL:} We also performed RL on instruction-tuned models that already possessing math reasoning ability. We explored this using R1-Distill-Qwen-1.5B~\citep{guo2025deepseek} and OpenMath-Nemotron-1.5B~\citep{moshkov2025aimo2}.
\end{itemize}

\paragraph{Evaluation} Following~\citet{zeng2025simplerlzooinvestigatingtamingzero,zeng2025simplerl}, we assessed the mathematical performance of our models on: MATH-500 \citep{hendrycksmath2021}, AMC 2023 \citep{amc}, OlympiadBench \citep{he2024olympiadbench}, Minerva Math \citep{minerva}, AIME 2024-2025 \citep{aime}, and the English subset of PolyMath~\citep{wang2025polymath}.
To investigate the generalization of reasoning abilities beyond mathematics, we used the GPQA-Diamond benchmark, which covers biology, physics and chemistry~\citep{gpqa}.
For all evaluations, we adopted pass@1 accuracy (averaged over 16 samples) as the metric, and fixed the decoding parameters to temperature=0.6, top\_p=0.95, and max\_tokens=32K.
To ensure a fair comparison and eliminate variance caused by the evaluation script, we re-evaluated the performance of all baseline models under our evaluation settings.

\subsection{Mathematical Reasoning Results}
\begin{table}[t]
    \centering
    \caption{Math reasoning performance. \myhl[tblue]{``DeepMath''} denotes models trained on \method{}.}
    \resizebox{1.0\linewidth}{!}{
    \begin{tabular}{l rrrrrrr}
    \toprule
    \multirow{2}{*}{\bf Model} &\bf MATH &\bf AMC &\bf Olympiad&\bf Minerva &\bf AIME &\bf AIME&\bf Poly\\
    &\bf 500  &\bf 23 &\bf Bench &\bf Math & \bf 24 & \bf 25& \bf Math\\
    \midrule
    \multicolumn{7}{c}{\textbf{\textit{Proprietary Models}}} \\
    o1-mini                                                                             &     --&     --&     --&     --&   63.6&     -- &    -- \\
    o3-mini (low effort)                                                                &     --&     --&     --&     --&   60.0&     -- &    -- \\
    \midrule
    \multicolumn{7}{c}{\textbf{\textit{Zero RL from Base Model}}} \\
    Qwen-2.5-7B~\citep{qwen2.5}                                                         &   54.8&   35.3&   27.8&   16.2&   7.7 &   5.4  &   28.1\\
    $\drsh$ Open-Reasoner-Zero-7B~\citep{OpenReasonerZero2025}                          &   81.8&   58.9&   47.9&   38.4&   15.6&   14.4 &   40.7\\
    $\drsh$ Qwen-2.5-7B-SRL-Zoo~\citep{zeng2025simplerlzooinvestigatingtamingzero}      &   77.0&   55.8&   41.0&   41.2&   15.6&   8.7  &   33.1\\
    \rowcolor{tblue}$\drsh$ DeepMath-Zero-7B (Ours)                                     &\bf85.5&\bf64.7&\bf51.0&\bf45.3&\bf20.4&\bf17.5 &\bf42.7\\
    \noalign{\medskip}
    Qwen-2.5-Math-7B~\citep{qwen2.5}                                                    &   46.9&   31.9&   15.8&   15.5&   11.2&   4.4  &   22.7\\
    $\drsh$ Qwen-2.5-Math-7B-SRL-Zoo~\citep{OpenReasonerZero2025}                       &   75.8&   59.7&   37.4&   29.9&   24.0&   10.2 &   36.0\\
    $\drsh$ Qat-Zero-7B~\citep{liu2025understanding}                                    &   80.0&   66.7&   43.4&   40.8&   32.7&   11.7 &   40.8\\
    $\drsh$ Eurus-2-7B-PRIME~\citep{cui2025process}                                     &   80.2&   64.7&   44.9&   42.1&   19.0&   12.7 &   38.9\\
    \rowcolor{tblue}$\drsh$ DeepMath-Zero-Math-7B (Ours)                                &\bf86.9&\bf74.7&\bf52.3&\bf49.5&\bf34.2&\bf23.5 &\bf46.6\\
    \midrule
    \multicolumn{7}{c}{\textbf{\textit{RL from Instruct Models}}} \\
    R1-Distill-Qwen-1.5B~\citep{guo2025deepseek}                                        &   84.7&   72.0&   53.1&   36.6&   29.4&   24.8 &   39.9\\
    $\drsh$ DeepScaleR-1.5B-Preview~\citep{deepscaler2025}                              &   89.4&   80.3&   60.9&   42.2&\bf42.3&   29.6 &\bf46.8\\
    $\drsh$ Still-3-1.5B-Preview~\citep{sitll3_1}                                       &   86.6&   75.8&   55.7&   38.7&   30.8&   24.6 &   43.1\\
    \rowcolor{tblue}$\drsh$ DeepMath-1.5B (Ours)                                        &\bf89.9&\bf82.3&\bf61.8&\bf42.5&   37.3&\bf30.8 &   46.6\\
    \noalign{\medskip}
    OpenMath-Nemotron-1.5B~\citep{moshkov2025aimo2}                                     &   91.8&   90.5&   70.3&   26.3&   61.3&   50.6 &   56.8\\
    \rowcolor{tblue}$\drsh$ DeepMath-Omn-1.5B (Ours)                                    &\bf93.2&\bf94.2&\bf73.4&\bf28.3&\bf64.0&\bf57.3 &\bf58.7\\
    \bottomrule
    \end{tabular}}
    \label{tab:main-result}
\end{table}
The results presented in \Cref{tab:main-result} collectively demonstrate the effectiveness of \method{} as a valuable resource for advancing the state-of-the-art in mathematical reasoning:

\paragraph{Zero RL Training on Base Model} DeepMath-Zero-7B and DeepMath-Zero-Math-7B, trained from the base Qwen-2.5-7B and Qwen-2.5-Math-7B models, demonstrate significant performance gains and achieve new \textbf{SOTA} results on all evaluated benchmarks.
These results highlight the effectiveness of \method{} in enabling the training of powerful mathematical reasoners from scratch.

\paragraph{RL Training on Instruction-tuned Models} Fine-tuning instruction-tuned models with RLVR on \method{} also yields notable performance enhancements. DeepMath-1.5B, initialized from R1-Distill-Qwen-1.5B, achieves strong performance, particularly on AMC23 (82.3\%) and OlympiadBench (61.8\%).
Similarly, DeepMath-Omn-1.5B, starting from OpenMath-Nemotron-1.5B, attains new \textbf{SOTA} results among 1.5B-scale models on all evaluated benchmarks, and even surpasses \textbf{o1-mini} and \textbf{o3-mini (low effort)}.
The consistent improvements observed across different instruction-tuned baselines further validate the utility of \method{} in boosting strong models.

\subsection{Generalizable Reasoning beyond Mathematics}
\Cref{tab:gpqa-diamond} presents the reasoning performance of DeepMath models on the GPQA-Diamond~\citep{gpqa}, which covers biology, physics, and chemistry.
DeepMath series models achieve superior performance compared to other baseline, demonstrating a remarkable capacity to generalize their reasoning abilities acquired from mathematics to broader domains.
We attribute this generalization to the data diversity and rigorous curation of \method{}.
By sourcing less structured but more diverse data like Math StackExchange, \method{} yields a dataset with unique and diverse problems.
Furthermore, the rigorous curation pipeline ensures both the challenge and the integrity of the data.
This exposure to a wider variety of problem-solving scenarios and reasoning styles likely equips our models with more robust and transferable reasoning skills.
\begin{table}[htpb]
    \centering
    \caption{Performance on the GPQA-Diamond benchmark.}
    \resizebox{0.75\linewidth}{!}{
    \begin{tabular}{l rrrr}
    \toprule
    \bf Model &\bf Biology &\bf Physics &\bf Chemistry &\bf Overall\\
    \midrule
    \multicolumn{5}{c}{\textbf{\textit{Zero RL from Base Model}}} \\
    Qwen-2.5-7B                                          &   33.6&   27.8&   21.4&   25.3\\
    $\drsh$ Open-Reasoner-Zero-7B                        &   50.3&   47.8&   26.7&   38.1\\
    $\drsh$ Qwen-2.5-7B-SimpleRL-Zoo                     &   31.9&   37.9&   22.6&   30.2\\
    \rowcolor{tblue}$\drsh$ DeepMath-Zero-7B (Ours)      &\bf57.2&\bf53.0&\bf28.2&\bf41.7\\
    \noalign{\medskip}
    Qwen-2.5-Math-7B                                     &   32.2&   26.0&   21.1&   24.3\\
    $\drsh$ Qwen-2.5-Math-7B-SRL-Zoo                     &   40.1&   31.2&   22.9&   28.2\\
    $\drsh$ Qat-Zero-7B                                  &   49.0&   36.8&   22.0&   31.0\\
    $\drsh$ Eurus-2-7B-PRIME                             &   44.1&   37.4&   24.1&   31.8\\
    \rowcolor{tblue}$\drsh$ DeepMath-Zero-Math-7B (Ours) &\bf47.4&\bf56.3&\bf26.0&\bf41.2\\
    \midrule
    \multicolumn{5}{c}{\textbf{\textit{RL from Instruct Models}}} \\
    R1-Distill-Qwen-1.5B                                 &   13.5&   36.2&    4.4&   19.1\\
    $\drsh$ DeepScaleR-1.5B-Preview                      &   15.5&   46.8&    9.1&   26.1\\
    $\drsh$ Still-3-1.5B-Preview                         &   16.8&   38.4&    5.2&   20.7\\
    \rowcolor{tblue}$\drsh$ DeepMath-1.5B (Ours)         &\bf18.1&\bf47.6&\bf12.2&\bf28.2\\
    \noalign{\medskip}
    OpenMath-Nemotron-1.5B                               &   12.8&   23.5&   18.9&   20.3\\
    \rowcolor{tblue}$\drsh$ DeepMath-Omn-1.5B (Ours)     &\bf17.1&\bf28.4&\bf21.5&\bf24.1\\
    \bottomrule
    \end{tabular}}
    \label{tab:gpqa-diamond}
\end{table}

\subsection{Analysis of Zero RL Using DeepMath-103K}
\begin{figure}[htpb]
    \begingroup
    \centering
    \setlength{\commoncontentheight}{4.5cm}
    \subfloat[Response length (training)\label{fig:zero-rl-train-len}]{
        \begin{minipage}[t][\commoncontentheight][t]{0.315\linewidth}
        \adjustimage{width=\linewidth, valign=t}{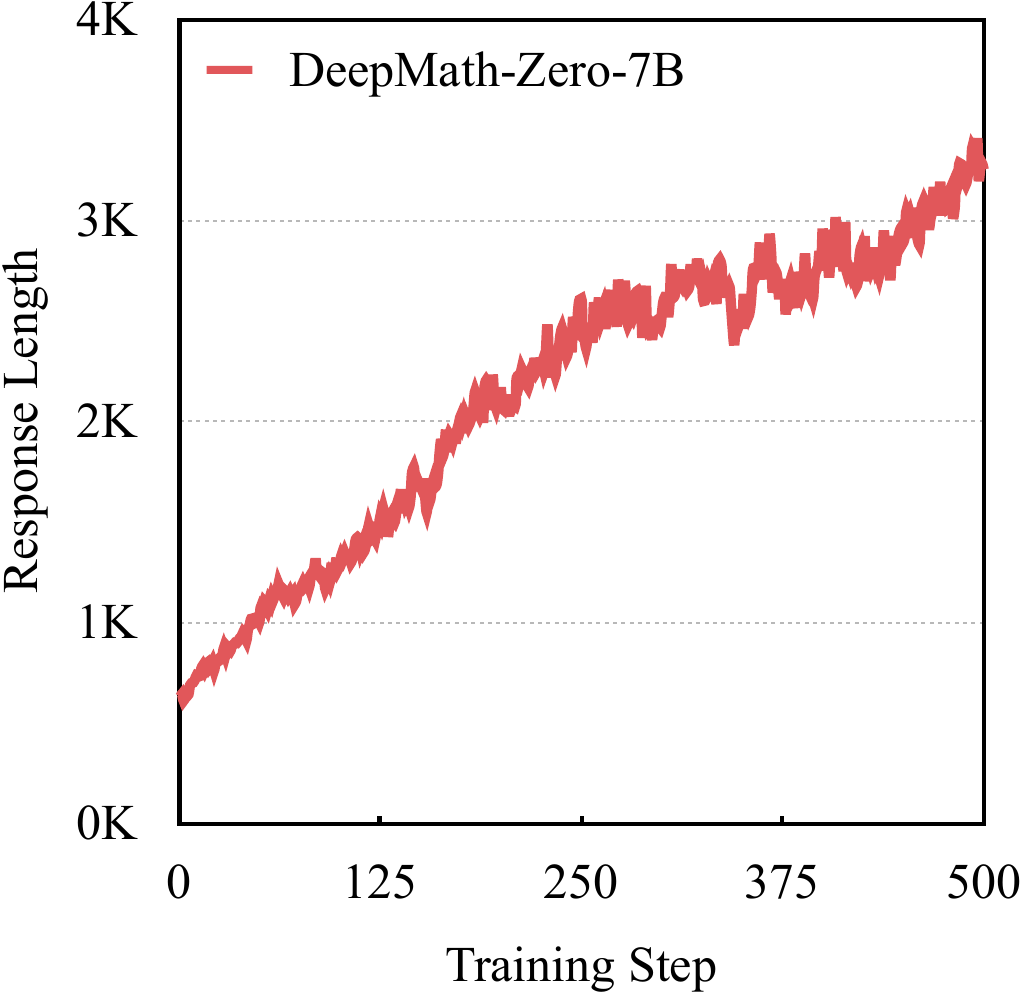}
        \end{minipage}
    }
    \hfill
    \subfloat[Change of cognitive behaviors.\label{fig:zero-rl-behavior}]{
        \begin{minipage}[t][\commoncontentheight][t]{0.325\linewidth}
        \adjustimage{width=\linewidth, valign=t}{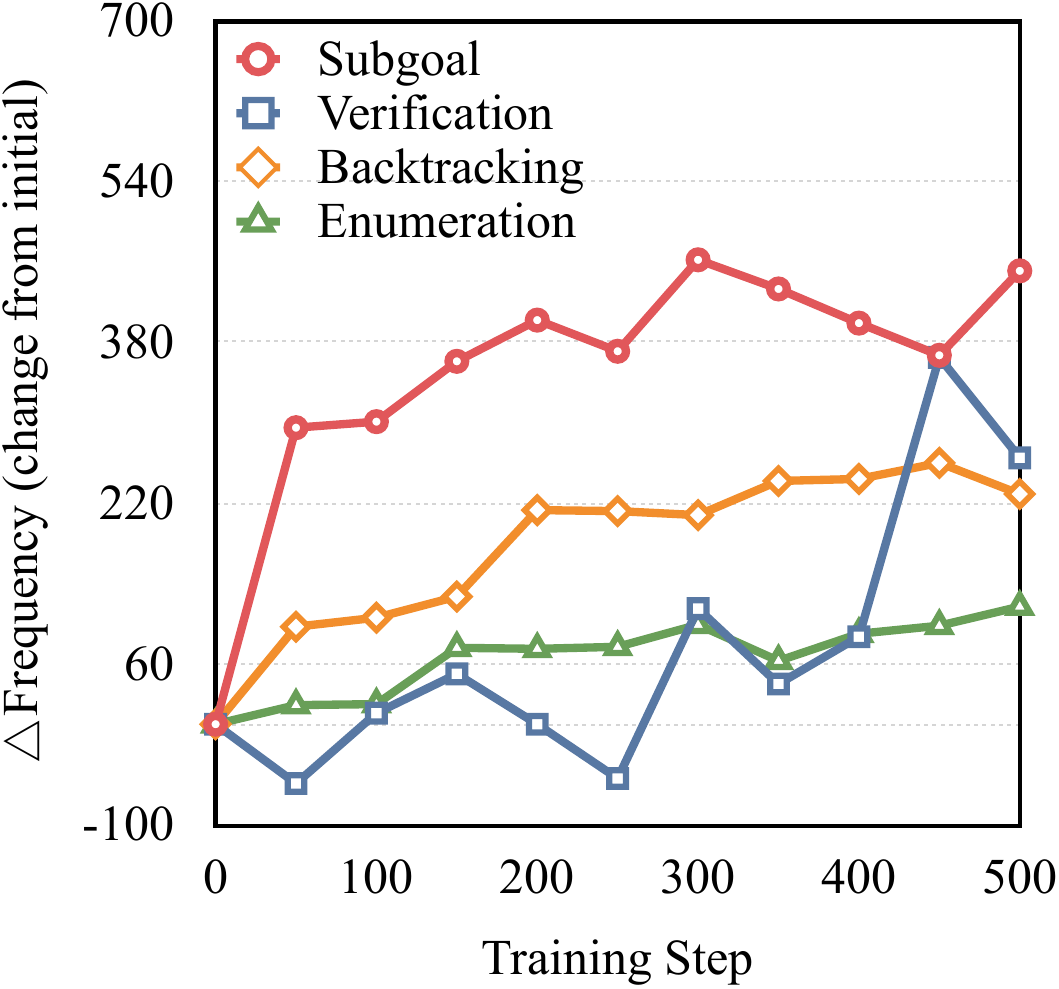}
        \end{minipage}
    }
    \hfill
    \subfloat[Response length (test)\label{fig:zero-rl-eval-len}]{
        \begin{minipage}[t][\commoncontentheight][t]{0.307\linewidth}
        \adjustimage{width=\linewidth, valign=t}{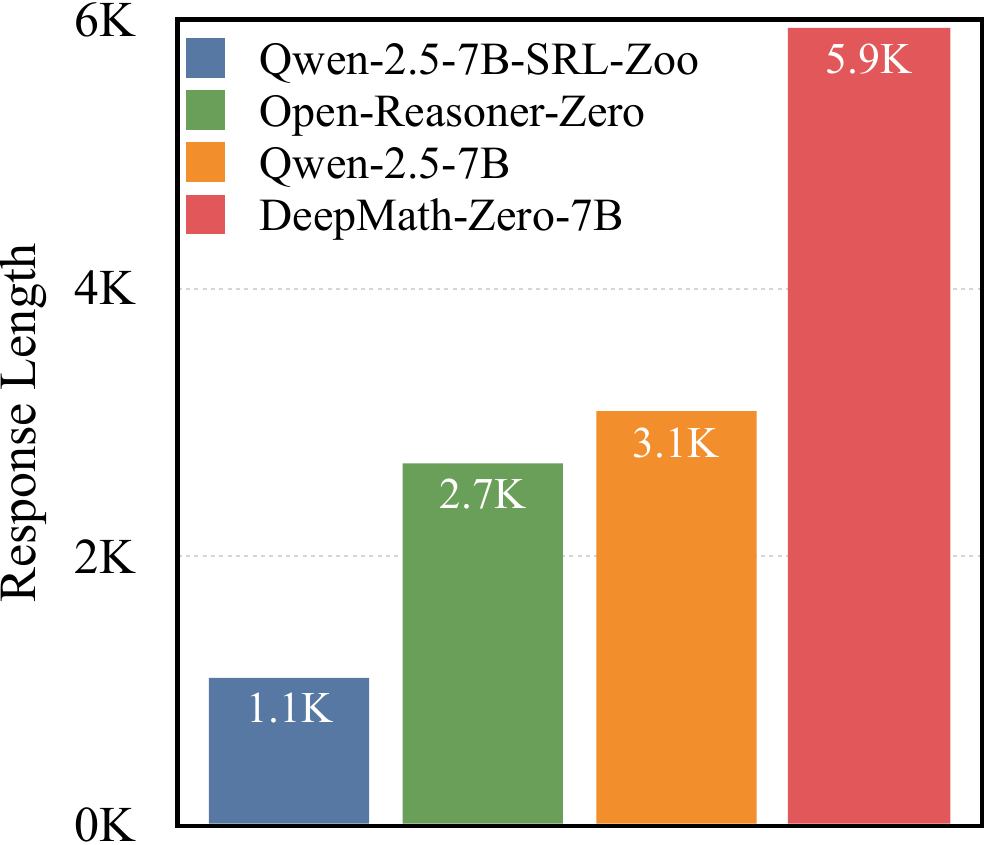}
        \end{minipage}
    }
    \caption{(a) \method{} is challenging compared to existing datasets. (b) Results of zero RL and RL using \method{} on AIME25.} 
    \label{fig:zero-rl-analysis}
    \endgroup
\end{figure}

\Cref{fig:zero-rl-analysis} presents an analysis of the characteristics observed during the zero RL training of DeepMath-Zero-7B.
Specifically, \Cref{fig:zero-rl-train-len} illustrates the trend of rollout response length throughout the training process, while \Cref{fig:zero-rl-behavior} tracks the emergence of four cognitive behaviors following the method outlined by~\citet{gandhi2025cognitive} and~\citet{zeng2025simplerlzooinvestigatingtamingzero}.
The increasing trends in both response length and the manifestation of cognitive behaviors suggest a reproduction of the ``aha moment'' phenomenon observed in DeepSeek-R1~\citep{guo2025deepseek}, which leads to a long reasoning model.
Furthermore, \Cref{fig:zero-rl-eval-len} shows the average response lengths of different models on the evaluated benchmarks.
The notably longer response lengths exhibited by DeepMath-Zero-7B indicate that \method{} serves as a valuable resource for research on long reasoning models, particularly concerning phenomena such as over- and under-thinking~\citep{chen2025ot,wang2025ut}.
\section{Related Work}
Datasets for advancing mathematical reasoning of LLM falls into three main lines corresponding to the three stages of LLM post-training: continue pre-training (CPT), SFT and RL.
CPT aims to inject fundamental mathematical knowledge into LLMs with representative works like OpenWebMath~\citep{paster2023openwebmath}, MathPile~\citep{wang2024mathpile}, InfiMM-Web-Math~\citep{han2024infimmwebmath40badvancingmultimodalpretraining}, FineMath~\citep{allal2025smollm2smolgoesbig}, and MegaMath~\citep{zhou2025megamath}.
SFT has been a foundational approach, utilizing datasets like MATH \citep{hendrycksmath2021} and GSM8K \citep{cobbe2021gsm8k} which provide problems with step-by-step solutions to teach models reasoning patterns.
Subsequent efforts have focused on creating larger, harder and more diverse SFT datasets, such as MetaMathQA~\citep{yumetamath}, OpenMathInstruct~\citep{toshniwal2024openmath,toshniwal2024openmath2}, NuminaMath-CoT~\citep{numina_math_datasets}, MMIQC~\citep{liu2024augmentingmathwordproblems}, dart-math-hard~\citep{tong2024dartmath}, and OpenMathReasoning~\citep{moshkov2025aimo2}.
Recent progress in RLVR catalyzes datasets with verifiable reward, such as Open-R1~\citep{openr1}, ORZ-129K~\citep{OpenReasonerZero2025}, DSR-Preview~\citep{deepscaler2025}, Open-RS~\citep{dang2025reinforcementlearningreasoningsmall}, DAPO-17K~\citep{yu2025dapoopensourcellmreinforcement}, and BigMath~\citep{albalak2025bigmathlargescalehighqualitymath}.
\method{} distinguishes itself by a unique blend of high difficulty, rigorous decontamination, and verifiable answers.
\section{Conclusion}
In this work, we introduce \method{}, a large-scale mathematical dataset specifically designed to advance the reasoning capabilities of LLMs through RLVR.
\method{} distinguishes itself through its high concentration of challenging problems, rigorous decontamination against a wide range of benchmarks, and the inclusion of verifiable final answers and multiple diverse solutions for each problem.
Our data curation pipeline leverages the richness of less structured mathematical forums, resulting in a dataset with significant novelty and diversity compared to existing resources.
Our experiments demonstrate the substantial impact of \method{}.
Models trained on this dataset, the DeepMath series, achieve new SOTA results on many mathematical benchmarks and exhibit remarkable generalization to domains beyond mathematics.
By releasing the \method{} dataset, along with our code and model weights, we aim to provide a robust platform for the community to further explore and push the boundaries of advanced reasoning.

\appendix
\clearpage
\section{Contamination Analysis of Existing Datasets}
\label{app:contamination}
\begin{wrapfigure}{r}{0.5\linewidth}
    \vspace{-12pt}
    \centering
    \includegraphics[width=\linewidth]{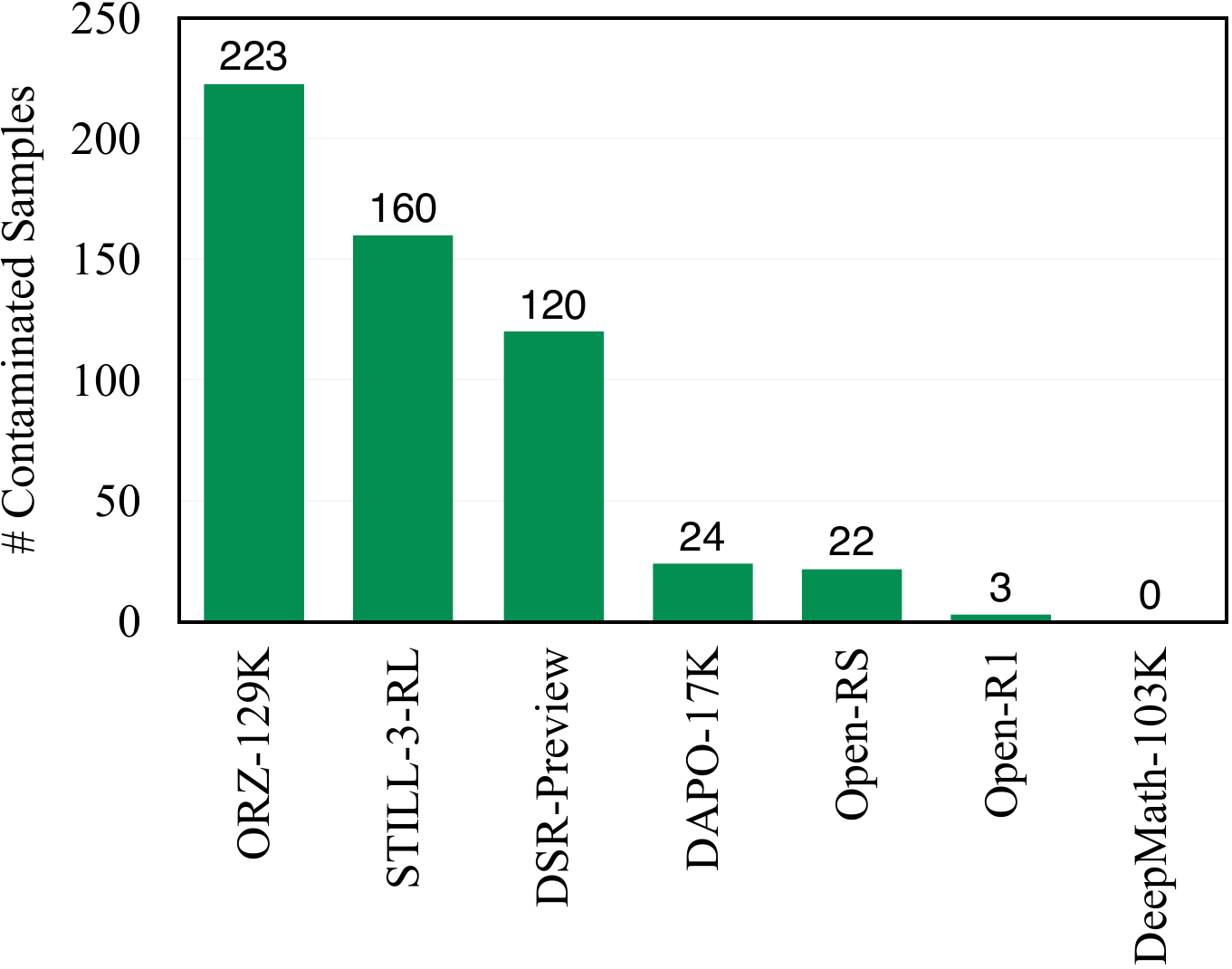}
    \caption{Number of contaminated samples in various datasets when compared against the MATH500 benchmark.}
    \label{fig:contamination-other}
    \vspace{-10pt}
\end{wrapfigure}
We performed a contamination analysis of several existing datasets, including ORZ-129K~\citep{OpenReasonerZero2025}, DSR-Preview~\citep{deepscaler2025}, DAPO-17K~\citep{yu2025dapoopensourcellmreinforcement}, Open-RS~\citep{bansal24smaller}, Open-R1~\citep{openr1}, and \method{}.
Our analysis focused on detecting potential contamination from the MATH500~\citep{hendrycksmath2021}, a commonly used benchmark.
We employed a string-based comparison method, specifically identifying cases where the normalized indel similarity between a problem in the analyzed dataset and a problem in MATH500 exceeded 90\%.
This approach is notably more lenient than the rigorous semantic decontamination procedure used in the construction of \method{} (\Cref{sec:curation}).
However, the numbers of contaminated samples shown in~\Cref{fig:contamination-other} reveal that most of the analyzed datasets exhibit some degree of contamination, with the exception of \method{}.

\section{SFT Results}
\label{sec:sft-results}
As mentioned in~\Cref{sec:overview}, each problem in \method{} includes three distinct R1-generated solutions, facilitating SFT.
We fine-tuned Qwen-2.5-7B using either the first R1 solution or all three solutions.
\Cref{tab:sft-result} shows that SFT on DeepMath-103K with one solution each problem also significantly enhances base model performance across all benchmarks, with multiple solutions yielding further gains.
However, SFT still lags behind RL.
\begin{table}[htpb]
    \centering
    \caption{Math reasoning performance after fine-tuning Qwen-2.5-7B via SFT. We also add DeepMath-Zero-7B as an RL counterpart for reference.}
    \resizebox{1.0\linewidth}{!}{
    \begin{tabular}{l rrrrrr}
    \toprule
    \multirow{2}{*}{\bf Model} &\multirow{2}{*}{\bf MATH500}  &\multirow{2}{*}{\bf AMC23} &\bf Olympiad&\bf Minerva &\multirow{2}{*}{\bf AIME24} &\multirow{2}{*}{\bf AIME25}\\
    & & &\bf Bench &\bf Math & &\\
    \midrule
    Qwen-2.5-7B                    &   54.8&   35.3&   27.8&   16.2&   7.7 &   5.4  \\
    $\drsh$ SFT with 1 R1 Solution &   69.2&   47.3&   35.9&   29.8&   12.3&   8.7  \\
    $\drsh$ SFT with 3 R1 Solutions&   74.1&   50.0&   40.2&   34.1&   13.8&  14.0  \\
    $\drsh$ DeepMath-Zero-7B (RL)  &\bf85.5&\bf64.7&\bf51.0&\bf45.3&\bf20.4&\bf17.5 \\
    \bottomrule
    \end{tabular}}
    \label{tab:sft-result}
\end{table}

\newpage
\section{Training Details}
We use \texttt{verl} as the training framework\footnote{\url{https://github.com/volcengine/verl}}.
Configurations for training DeepMath series models are listed in~\Cref{tab:training-detail}.
\label{app:training-detail}
\begin{table}[htpb]
    \centering
    \caption{Configurations for training DeepMath series models.} 
    \resizebox{\linewidth}{!}{
    \begin{tabular}{l l l l l l l l l l l}
    \toprule
     Config & DeepMath-Zero-7B & DeepMath-Zero-Math-7B & DeepMath-1.5B & DeepMath-Omn-1.5B \\
    \midrule
    lr                     & 1e-6 & 1e-6 & 1e-6 & 1e-6\\
    kl\_coef               & 0.0  & 0.0  & 1e-3 & 1e-3\\
    max\_prompt\_length    & 2K   & 1K   & 2K   & 2K  \\
    max\_response\_length  & 10K  & 3K   & 24K  & 24K \\
    train\_batch\_size     & 512  & 512  & 128  & 128 \\
    ppo\_mini\_batch\_size & 32   & 32   & 64   & 64  \\
    clip\_ratio\_low       & 0.20 & 0.20 & 0.20 & 0.20\\
    clip\_ratio\_high      & 0.28 & 0.28 & 0.27 & 0.27\\
    temperature            & 1.0  & 1.0  & 0.6  & 0.6 \\
    rollout.n              & 16   & 16   & 16   & 18  \\
    overlong\_buffer.len   & 2K   & 512  & 4K   & 4K  \\
    total\_training\_steps & 500  & 500  & 1800 & 700 \\
    \bottomrule
    \end{tabular}}
    \label{tab:training-detail}
\end{table}

\section{Licenses for Existing Assets}
\label{app:licenses-for-existing-assets}
\begin{table}[htpb]
    \centering
    \caption{Licenses for existing assets.}
    \begin{tabular}{l l l}
    \toprule
    \bf Type & \bf Asset & \bf License \\
    \midrule
    \multirow{3}{*}{Data}  & MMIQC~\citep{liu2024augmentingmathwordproblems}     & Apache 2.0 \\
                           & WebInstSub~\citep{yue2024mammoth2}                  & Apache 2.0 \\
                           & NuminaMath-CoT~\citep{numina_math_datasets}         & Apache 2.0 \\
    \midrule
    \multirow{3}{*}{Code}  & verl~\citep{sheng2024hybridflow}                    & Apache 2.0 \\
                           & NeMo-Skills~\citep{toshniwal2024openmath}  & Apache 2.0 \\
    \midrule
    \multirow{3}{*}{Model} & Qwen-2.5-7B~\cite{qwen2.5}                          & Apache 2.0 \\
                           & R1-Distill-Qwen-1.5B~\citep{guo2025deepseek}        & MIT        \\
                           & OpenMath-Nemotron-1.5B~\citep{moshkov2025aimo2}     & CC BY 4.0  \\
    \bottomrule
    \end{tabular}
    \label{tab:licenses-for-existing-assets}
\end{table}

\section{Limitations and Broader Impacts}
\label{app:limitations-and-impact}
While \method{} advances mathematical AI, its difficulty assessment relies on LLM evaluations, potentially introducing bias.
Topical diversity may not be perfectly balanced, and the dataset's creation was computationally intensive.
Our manual analysis reveals judgment and multiple-choice questions whose answers might be matched successfully via random guess.
However, \method{}'s public release can lower the barrier for RL reasoning research, accelerate progress on challenging problems, improve benchmark reliability, and foster more generalizable AI.

\bibliography{ref}

\begin{thebibliography}{49}
\providecommand{\natexlab}[1]{#1}
\providecommand{\url}[1]{\texttt{#1}}
\expandafter\ifx\csname urlstyle\endcsname\relax
  \providecommand{\doi}[1]{doi: #1}\else
  \providecommand{\doi}{doi: \begingroup \urlstyle{rm}\Url}\fi

\bibitem[Albalak et~al.(2025)Albalak, Phung, Lile, Rafailov, Gandhi, Castricato, Singh, Blagden, Xiang, Mahan, and Haber]{albalak2025bigmathlargescalehighqualitymath}
Alon Albalak, Duy Phung, Nathan Lile, Rafael Rafailov, Kanishk Gandhi, Louis Castricato, Anikait Singh, Chase Blagden, Violet Xiang, Dakota Mahan, and Nick Haber.
\newblock Big-math: A large-scale, high-quality math dataset for reinforcement learning in language models, 2025.
\newblock URL \url{https://arxiv.org/abs/2502.17387}.

\bibitem[Allal et~al.(2025)Allal, Lozhkov, Bakouch, Blázquez, Penedo, Tunstall, Marafioti, Kydlíček, Lajarín, Srivastav, Lochner, Fahlgren, Nguyen, Fourrier, Burtenshaw, Larcher, Zhao, Zakka, Morlon, Raffel, von Werra, and Wolf]{allal2025smollm2smolgoesbig}
Loubna~Ben Allal, Anton Lozhkov, Elie Bakouch, Gabriel~Martín Blázquez, Guilherme Penedo, Lewis Tunstall, Andrés Marafioti, Hynek Kydlíček, Agustín~Piqueres Lajarín, Vaibhav Srivastav, Joshua Lochner, Caleb Fahlgren, Xuan-Son Nguyen, Clémentine Fourrier, Ben Burtenshaw, Hugo Larcher, Haojun Zhao, Cyril Zakka, Mathieu Morlon, Colin Raffel, Leandro von Werra, and Thomas Wolf.
\newblock Smollm2: When smol goes big -- data-centric training of a small language model, 2025.
\newblock URL \url{https://arxiv.org/abs/2502.02737}.

\bibitem[Arora et~al.(2023)Arora, Singh, and {Mausam}]{arora-etal-2023-llms}
Daman Arora, Himanshu Singh, and {Mausam}.
\newblock Have {LLM}s advanced enough? a challenging problem solving benchmark for large language models.
\newblock In Houda Bouamor, Juan Pino, and Kalika Bali (eds.), \emph{Proceedings of the 2023 Conference on Empirical Methods in Natural Language Processing}, pp.\  7527--7543, Singapore, December 2023. Association for Computational Linguistics.
\newblock \doi{10.18653/v1/2023.emnlp-main.468}.
\newblock URL \url{https://aclanthology.org/2023.emnlp-main.468}.

\bibitem[Bansal et~al.(2025)Bansal, Hosseini, Agarwal, Tran, and Kazemi]{bansal24smaller}
Hritik Bansal, Arian Hosseini, Rishabh Agarwal, Vinh~Q. Tran, and Mehran Kazemi.
\newblock Smaller, weaker, yet better: Training {LLM} reasoners via compute-optimal sampling.
\newblock In \emph{The Thirteenth International Conference on Learning Representations}, 2025.
\newblock URL \url{https://openreview.net/forum?id=3OyaXFQuDl}.

\bibitem[Chen et~al.(2024)Chen, Xu, Liang, He, Pang, Yu, Song, Liu, Zhou, Zhang, Wang, Tu, Mi, and Yu]{chen2025ot}
Xingyu Chen, Jiahao Xu, Tian Liang, Zhiwei He, Jianhui Pang, Dian Yu, Linfeng Song, Qiuzhi Liu, Mengfei Zhou, Zhuosheng Zhang, Rui Wang, Zhaopeng Tu, Haitao Mi, and Dong Yu.
\newblock Do not think that much for 2+3=? on the overthinking of o1-like llms, 2024.
\newblock URL \url{https://arxiv.org/abs/2412.21187}.

\bibitem[Chen et~al.(2025)Chen, Min, Zhang, Chen, Jiang, Cheng, Zhao, Liu, Miao, Lu, Fang, Wang, and Wen]{sitll3_1}
Zhipeng Chen, Yingqian Min, Beichen Zhang, Jie Chen, Jinhao Jiang, Daixuan Cheng, Wayne~Xin Zhao, Zheng Liu, Xu~Miao, Yang Lu, Lei Fang, Zhongyuan Wang, and Ji-Rong Wen.
\newblock An empirical study on eliciting and improving r1-like reasoning models.
\newblock \emph{arXiv preprint arXiv:2503.04548}, 2025.

\bibitem[Cobbe et~al.(2021)Cobbe, Kosaraju, Bavarian, Chen, Jun, Kaiser, Plappert, Tworek, Hilton, Nakano, Hesse, and Schulman]{cobbe2021gsm8k}
Karl Cobbe, Vineet Kosaraju, Mohammad Bavarian, Mark Chen, Heewoo Jun, Lukasz Kaiser, Matthias Plappert, Jerry Tworek, Jacob Hilton, Reiichiro Nakano, Christopher Hesse, and John Schulman.
\newblock Training verifiers to solve math word problems.
\newblock \emph{arXiv preprint arXiv:2110.14168}, 2021.

\bibitem[Cui et~al.(2025)Cui, Yuan, Wang, Wang, Li, He, Fan, Yu, Xu, Chen, et~al.]{cui2025process}
Ganqu Cui, Lifan Yuan, Zefan Wang, Hanbin Wang, Wendi Li, Bingxiang He, Yuchen Fan, Tianyu Yu, Qixin Xu, Weize Chen, et~al.
\newblock Process reinforcement through implicit rewards.
\newblock \emph{arXiv preprint arXiv:2502.01456}, 2025.

\bibitem[Dang \& Ngo(2025)Dang and Ngo]{dang2025reinforcementlearningreasoningsmall}
Quy-Anh Dang and Chris Ngo.
\newblock Reinforcement learning for reasoning in small llms: What works and what doesn't, 2025.
\newblock URL \url{https://arxiv.org/abs/2503.16219}.

\bibitem[Face(2025)]{openr1}
Hugging Face.
\newblock Open r1: A fully open reproduction of deepseek-r1, January 2025.
\newblock URL \url{https://github.com/huggingface/open-r1}.

\bibitem[Fang et~al.(2024)Fang, Wan, Lu, Xing, and Zou]{fang2024mathodyssey}
Meng Fang, Xiangpeng Wan, Fei Lu, Fei Xing, and Kai Zou.
\newblock Mathodyssey: Benchmarking mathematical problem-solving skills in large language models using odyssey math data.
\newblock \emph{arXiv preprint arXiv:2406.18321}, 2024.

\bibitem[Gandhi et~al.(2025)Gandhi, Chakravarthy, Singh, Lile, and Goodman]{gandhi2025cognitive}
Kanishk Gandhi, Ayush Chakravarthy, Anikait Singh, Nathan Lile, and Noah~D Goodman.
\newblock Cognitive behaviors that enable self-improving reasoners, or, four habits of highly effective stars.
\newblock \emph{arXiv preprint arXiv:2503.01307}, 2025.

\bibitem[Gao et~al.(2024)Gao, Song, Yang, Cai, Miao, Dong, Li, Ma, Chen, Xu, Tang, Wang, Zan, Quan, Zhang, Sha, Zhang, Ren, Liu, and Chang]{gao2024omnimathuniversalolympiadlevel}
Bofei Gao, Feifan Song, Zhe Yang, Zefan Cai, Yibo Miao, Qingxiu Dong, Lei Li, Chenghao Ma, Liang Chen, Runxin Xu, Zhengyang Tang, Benyou Wang, Daoguang Zan, Shanghaoran Quan, Ge~Zhang, Lei Sha, Yichang Zhang, Xuancheng Ren, Tianyu Liu, and Baobao Chang.
\newblock Omni-math: A universal olympiad level mathematic benchmark for large language models, 2024.
\newblock URL \url{https://arxiv.org/abs/2410.07985}.

\bibitem[Grattafiori et~al.(2024)Grattafiori, Dubey, Jauhri, Pandey, Kadian, Al-Dahle, Letman, Mathur, Schelten, Vaughan, et~al.]{grattafiori2024llama}
Aaron Grattafiori, Abhimanyu Dubey, Abhinav Jauhri, Abhinav Pandey, Abhishek Kadian, Ahmad Al-Dahle, Aiesha Letman, Akhil Mathur, Alan Schelten, Alex Vaughan, et~al.
\newblock The llama 3 herd of models.
\newblock \emph{arXiv preprint arXiv:2407.21783}, 2024.

\bibitem[Guo et~al.(2025)Guo, Yang, Zhang, Song, Zhang, Xu, Zhu, Ma, Wang, Bi, et~al.]{guo2025deepseek}
Daya Guo, Dejian Yang, Haowei Zhang, Junxiao Song, Ruoyu Zhang, Runxin Xu, Qihao Zhu, Shirong Ma, Peiyi Wang, Xiao Bi, et~al.
\newblock Deepseek-r1: Incentivizing reasoning capability in llms via reinforcement learning.
\newblock \emph{arXiv preprint arXiv:2501.12948}, 2025.

\bibitem[Han et~al.(2024)Han, Jian, Hu, Liu, Wang, Fan, Ai, Huang, He, Yang, and You]{han2024infimmwebmath40badvancingmultimodalpretraining}
Xiaotian Han, Yiren Jian, Xuefeng Hu, Haogeng Liu, Yiqi Wang, Qihang Fan, Yuang Ai, Huaibo Huang, Ran He, Zhenheng Yang, and Quanzeng You.
\newblock Infimm-webmath-40b: Advancing multimodal pre-training for enhanced mathematical reasoning, 2024.
\newblock URL \url{https://arxiv.org/abs/2409.12568}.

\bibitem[He et~al.(2024)He, Luo, Bai, Hu, Thai, Shen, Hu, Han, Huang, Zhang, Liu, Qi, Liu, and Sun]{he2024olympiadbench}
Chaoqun He, Renjie Luo, Yuzhuo Bai, Shengding Hu, Zhen~Leng Thai, Junhao Shen, Jinyi Hu, Xu~Han, Yujie Huang, Yuxiang Zhang, Jie Liu, Lei Qi, Zhiyuan Liu, and Maosong Sun.
\newblock Olympiadbench: A challenging benchmark for promoting agi with olympiad-level bilingual multimodal scientific problems, 2024.

\bibitem[Hendrycks et~al.(2021{\natexlab{a}})Hendrycks, Burns, Basart, Zou, Mazeika, Song, and Steinhardt]{hendryckstest2021}
Dan Hendrycks, Collin Burns, Steven Basart, Andy Zou, Mantas Mazeika, Dawn Song, and Jacob Steinhardt.
\newblock Measuring massive multitask language understanding.
\newblock \emph{Proceedings of the International Conference on Learning Representations (ICLR)}, 2021{\natexlab{a}}.

\bibitem[Hendrycks et~al.(2021{\natexlab{b}})Hendrycks, Burns, Kadavath, Arora, Basart, Tang, Song, and Steinhardt]{hendrycksmath2021}
Dan Hendrycks, Collin Burns, Saurav Kadavath, Akul Arora, Steven Basart, Eric Tang, Dawn Song, and Jacob Steinhardt.
\newblock Measuring mathematical problem solving with the math dataset.
\newblock \emph{NeurIPS}, 2021{\natexlab{b}}.

\bibitem[Hu et~al.(2025)Hu, Zhang, Han, Jiang, and Xiangyu~Zhang]{OpenReasonerZero2025}
Jingcheng Hu, Yinmin Zhang, Qi~Han, Daxin Jiang, and Heung-Yeung~Shum Xiangyu~Zhang.
\newblock Open-reasoner-zero: An open source approach to scaling reinforcement learning on the base model.
\newblock \url{https://github.com/Open-Reasoner-Zero/Open-Reasoner-Zero}, 2025.

\bibitem[Huang et~al.(2024)Huang, Wang, Xia, Li, Zou, Xu, Fan, Ye, Chern, Ye, Zhang, Yang, Wu, Wang, Sun, Xiao, Li, Zhou, Chern, Qin, Ma, Su, Liu, Zheng, Zhang, Lin, Qiao, and Liu]{huang2024olympicarena}
Zhen Huang, Zengzhi Wang, Shijie Xia, Xuefeng Li, Haoyang Zou, Ruijie Xu, Run-Ze Fan, Lyumanshan Ye, Ethan Chern, Yixin Ye, Yikai Zhang, Yuqing Yang, Ting Wu, Binjie Wang, Shichao Sun, Yang Xiao, Yiyuan Li, Fan Zhou, Steffi Chern, Yiwei Qin, Yan Ma, Jiadi Su, Yixiu Liu, Yuxiang Zheng, Shaoting Zhang, Dahua Lin, Yu~Qiao, and Pengfei Liu.
\newblock Olympicarena: Benchmarking multi-discipline cognitive reasoning for superintelligent ai.
\newblock \emph{arXiv preprint arXiv:2406.12753}, 2024.
\newblock URL \url{https://arxiv.org/abs/2406.12753}.

\bibitem[Lewkowycz et~al.(2022)Lewkowycz, Andreassen, Dohan, Dyer, Michalewski, Ramasesh, Slone, Anil, Schlag, Gutman-Solo, Wu, Neyshabur, Gur-Ari, and Misra]{minerva}
Aitor Lewkowycz, Anders Andreassen, David Dohan, Ethan Dyer, Henryk Michalewski, Vinay Ramasesh, Ambrose Slone, Cem Anil, Imanol Schlag, Theo Gutman-Solo, Yuhuai Wu, Behnam Neyshabur, Guy Gur-Ari, and Vedant Misra.
\newblock Solving quantitative reasoning problems with language models.
\newblock In S.~Koyejo, S.~Mohamed, A.~Agarwal, D.~Belgrave, K.~Cho, and A.~Oh (eds.), \emph{Advances in Neural Information Processing Systems}, volume~35, pp.\  3843--3857. Curran Associates, Inc., 2022.
\newblock URL \url{https://proceedings.neurips.cc/paper_files/paper/2022/file/18abbeef8cfe9203fdf9053c9c4fe191-Paper-Conference.pdf}.

\bibitem[LI et~al.(2024)LI, Beeching, Tunstall, Lipkin, Soletskyi, Huang, Rasul, Yu, Jiang, Shen, Qin, Dong, Zhou, Fleureau, Lample, and Polu]{numina_math_datasets}
Jia LI, Edward Beeching, Lewis Tunstall, Ben Lipkin, Roman Soletskyi, Shengyi~Costa Huang, Kashif Rasul, Longhui Yu, Albert Jiang, Ziju Shen, Zihan Qin, Bin Dong, Li~Zhou, Yann Fleureau, Guillaume Lample, and Stanislas Polu.
\newblock Numinamath.
\newblock \url{[https://huggingface.co/AI-MO/NuminaMath-CoT](https://github.com/project-numina/aimo-progress-prize/blob/main/report/numina_dataset.pdf)}, 2024.

\bibitem[Liu et~al.(2024)Liu, Zhang, Luo, and Yao]{liu2024augmentingmathwordproblems}
Haoxiong Liu, Yifan Zhang, Yifan Luo, and Andrew Chi-Chih Yao.
\newblock Augmenting math word problems via iterative question composing, 2024.
\newblock URL \url{https://arxiv.org/abs/2401.09003}.

\bibitem[Liu et~al.(2025)Liu, Chen, Li, Qi, Pang, Du, Lee, and Lin]{liu2025understanding}
Zichen Liu, Changyu Chen, Wenjun Li, Penghui Qi, Tianyu Pang, Chao Du, Wee~Sun Lee, and Min Lin.
\newblock Understanding r1-zero-like training: A critical perspective.
\newblock \emph{arXiv preprint arXiv:2503.20783}, 2025.

\bibitem[Luo et~al.(2025)Luo, Tan, Wong, Shi, Tang, Roongta, Cai, Luo, Zhang, Li, Popa, and Stoica]{deepscaler2025}
Michael Luo, Sijun Tan, Justin Wong, Xiaoxiang Shi, William~Y. Tang, Manan Roongta, Colin Cai, Jeffrey Luo, Tianjun Zhang, Li~Erran Li, Raluca~Ada Popa, and Ion Stoica.
\newblock Deepscaler: Surpassing o1-preview with a 1.5b model by scaling rl, 2025.
\newblock Notion Blog.

\bibitem[MAA({\natexlab{a}})]{aime}
MAA.
\newblock American invitational mathematics examination ({AIME}).
\newblock Mathematics Competition Series, n.d.{\natexlab{a}}.
\newblock URL \url{https://maa.org/math-competitions/aime}.

\bibitem[MAA({\natexlab{b}})]{amc}
MAA.
\newblock American mathematics competitions ({AMC} 10/12).
\newblock Mathematics Competition Series, n.d.{\natexlab{b}}.
\newblock URL \url{https://maa.org/math-competitions/amc}.

\bibitem[Moshkov et~al.(2025)Moshkov, Hanley, Sorokin, Toshniwal, Henkel, Schifferer, Du, and Gitman]{moshkov2025aimo2}
Ivan Moshkov, Darragh Hanley, Ivan Sorokin, Shubham Toshniwal, Christof Henkel, Benedikt Schifferer, Wei Du, and Igor Gitman.
\newblock Aimo-2 winning solution: Building state-of-the-art mathematical reasoning models with openmathreasoning dataset.
\newblock \emph{arXiv preprint arXiv:2504.16891}, 2025.

\bibitem[Paster et~al.(2023)Paster, Santos, Azerbayev, and Ba]{paster2023openwebmath}
Keiran Paster, Marco~Dos Santos, Zhangir Azerbayev, and Jimmy Ba.
\newblock Openwebmath: An open dataset of high-quality mathematical web text, 2023.

\bibitem[Reimers \& Gurevych(2019)Reimers and Gurevych]{reimers-2019-sentence-bert}
Nils Reimers and Iryna Gurevych.
\newblock Sentence-bert: Sentence embeddings using siamese bert-networks.
\newblock In \emph{Proceedings of the 2019 Conference on Empirical Methods in Natural Language Processing}. Association for Computational Linguistics, 11 2019.
\newblock URL \url{http://arxiv.org/abs/1908.10084}.

\bibitem[Rein et~al.(2024)Rein, Hou, Stickland, Petty, Pang, Dirani, Michael, and Bowman]{gpqa}
David Rein, Betty~Li Hou, Asa~Cooper Stickland, Jackson Petty, Richard~Yuanzhe Pang, Julien Dirani, Julian Michael, and Samuel~R. Bowman.
\newblock {GPQA}: A graduate-level google-proof q\&a benchmark.
\newblock In \emph{First Conference on Language Modeling}, 2024.
\newblock URL \url{https://openreview.net/forum?id=Ti67584b98}.

\bibitem[Shao et~al.(2024)Shao, Wang, Zhu, Xu, Song, Bi, Zhang, Zhang, Li, Wu, and Guo]{shao2024deepseekmathpushinglimitsmathematical}
Zhihong Shao, Peiyi Wang, Qihao Zhu, Runxin Xu, Junxiao Song, Xiao Bi, Haowei Zhang, Mingchuan Zhang, Y.~K. Li, Y.~Wu, and Daya Guo.
\newblock Deepseekmath: Pushing the limits of mathematical reasoning in open language models, 2024.
\newblock URL \url{https://arxiv.org/abs/2402.03300}.

\bibitem[Sheng et~al.(2024)Sheng, Zhang, Ye, Wu, Zhang, Zhang, Peng, Lin, and Wu]{sheng2024hybridflow}
Guangming Sheng, Chi Zhang, Zilingfeng Ye, Xibin Wu, Wang Zhang, Ru~Zhang, Yanghua Peng, Haibin Lin, and Chuan Wu.
\newblock Hybridflow: A flexible and efficient rlhf framework.
\newblock \emph{arXiv preprint arXiv: 2409.19256}, 2024.

\bibitem[Team(2024)]{qwen2.5}
Qwen Team.
\newblock Qwen2.5: A party of foundation models, September 2024.
\newblock URL \url{https://qwenlm.github.io/blog/qwen2.5/}.

\bibitem[Tong et~al.(2024)Tong, Zhang, Wang, Wu, and He]{tong2024dartmath}
Yuxuan Tong, Xiwen Zhang, Rui Wang, Ruidong Wu, and Junxian He.
\newblock {DART}-math: Difficulty-aware rejection tuning for mathematical problem-solving.
\newblock In \emph{The Thirty-eighth Annual Conference on Neural Information Processing Systems}, 2024.
\newblock URL \url{https://openreview.net/forum?id=zLU21oQjD5}.

\bibitem[Toshniwal et~al.(2024{\natexlab{a}})Toshniwal, Du, Moshkov, Kisacanin, Ayrapetyan, and Gitman]{toshniwal2024openmath2}
Shubham Toshniwal, Wei Du, Ivan Moshkov, Branislav Kisacanin, Alexan Ayrapetyan, and Igor Gitman.
\newblock Openmathinstruct-2: Accelerating ai for math with massive open-source instruction data.
\newblock \emph{arXiv preprint arXiv:2410.01560}, 2024{\natexlab{a}}.

\bibitem[Toshniwal et~al.(2024{\natexlab{b}})Toshniwal, Moshkov, Narenthiran, Gitman, Jia, and Gitman]{toshniwal2024openmath}
Shubham Toshniwal, Ivan Moshkov, Sean Narenthiran, Daria Gitman, Fei Jia, and Igor Gitman.
\newblock Openmathinstruct-1: A 1.8 million math instruction tuning dataset.
\newblock \emph{arXiv preprint arXiv: Arxiv-2402.10176}, 2024{\natexlab{b}}.

\bibitem[Wang et~al.(2025{\natexlab{a}})Wang, Zhang, Tang, Wei, Yang, Wang, Sun, Sun, Zhang, Wu, Cang, Zhang, Huang, Lin, Huang, and Zhou]{wang2025polymath}
Yiming Wang, Pei Zhang, Jialong Tang, Haoran Wei, Baosong Yang, Rui Wang, Chenshu Sun, Feitong Sun, Jiran Zhang, Junxuan Wu, Qiqian Cang, Yichang Zhang, Fei Huang, Junyang Lin, Fei Huang, and Jingren Zhou.
\newblock Polymath: Evaluating mathematical reasoning in multilingual contexts.
\newblock \emph{arXiv preprint arXiv:2504.18428}, 2025{\natexlab{a}}.
\newblock URL \url{https://arxiv.org/abs/2504.18428}.

\bibitem[Wang et~al.(2025{\natexlab{b}})Wang, Liu, Xu, Liang, Chen, He, Song, Yu, Li, Zhang, Wang, Tu, Mi, and Yu]{wang2025ut}
Yue Wang, Qiuzhi Liu, Jiahao Xu, Tian Liang, Xingyu Chen, Zhiwei He, Linfeng Song, Dian Yu, Juntao Li, Zhuosheng Zhang, Rui Wang, Zhaopeng Tu, Haitao Mi, and Dong Yu.
\newblock Thoughts are all over the place: On the underthinking of o1-like llms, 2025{\natexlab{b}}.
\newblock URL \url{https://arxiv.org/abs/2501.18585}.

\bibitem[Wang et~al.(2024)Wang, Li, Xia, and Liu]{wang2024mathpile}
Zengzhi Wang, Xuefeng Li, Rui Xia, and Pengfei Liu.
\newblock Mathpile: A billion-token-scale pretraining corpus for math.
\newblock In \emph{The Thirty-eight Conference on Neural Information Processing Systems Datasets and Benchmarks Track}, 2024.
\newblock URL \url{https://openreview.net/forum?id=RSvhU69sbG}.

\bibitem[Wei et~al.(2023)Wei, Luan, Liu, Dong, and Wang]{wei2023cmath}
Tianwen Wei, Jian Luan, Wei Liu, Shuang Dong, and Bin Wang.
\newblock Cmath: Can your language model pass chinese elementary school math test?, 2023.

\bibitem[Yu et~al.(2024)Yu, Jiang, Shi, YU, Liu, Zhang, Kwok, Li, Weller, and Liu]{yumetamath}
Longhui Yu, Weisen Jiang, Han Shi, Jincheng YU, Zhengying Liu, Yu~Zhang, James Kwok, Zhenguo Li, Adrian Weller, and Weiyang Liu.
\newblock Metamath: Bootstrap your own mathematical questions for large language models.
\newblock In \emph{The Twelfth International Conference on Learning Representations}, 2024.
\newblock URL \url{https://openreview.net/forum?id=N8N0hgNDRt}.

\bibitem[Yu et~al.(2025)Yu, Zhang, Zhu, Yuan, Zuo, Yue, Fan, Liu, Liu, Liu, Lin, Lin, Ma, Sheng, Tong, Zhang, Zhang, Zhang, Zhu, Zhu, Chen, Chen, Wang, Yu, Dai, Song, Wei, Zhou, Liu, Ma, Zhang, Yan, Qiao, Wu, and Wang]{yu2025dapoopensourcellmreinforcement}
Qiying Yu, Zheng Zhang, Ruofei Zhu, Yufeng Yuan, Xiaochen Zuo, Yu~Yue, Tiantian Fan, Gaohong Liu, Lingjun Liu, Xin Liu, Haibin Lin, Zhiqi Lin, Bole Ma, Guangming Sheng, Yuxuan Tong, Chi Zhang, Mofan Zhang, Wang Zhang, Hang Zhu, Jinhua Zhu, Jiaze Chen, Jiangjie Chen, Chengyi Wang, Hongli Yu, Weinan Dai, Yuxuan Song, Xiangpeng Wei, Hao Zhou, Jingjing Liu, Wei-Ying Ma, Ya-Qin Zhang, Lin Yan, Mu~Qiao, Yonghui Wu, and Mingxuan Wang.
\newblock Dapo: An open-source llm reinforcement learning system at scale, 2025.
\newblock URL \url{https://arxiv.org/abs/2503.14476}.

\bibitem[Yue et~al.(2024)Yue, Zheng, Zhang, and Chen]{yue2024mammoth2}
Xiang Yue, Tianyu Zheng, Ge~Zhang, and Wenhu Chen.
\newblock Mammoth2: Scaling instructions from the web.
\newblock \emph{Advances in Neural Information Processing Systems}, 37:\penalty0 90629--90660, 2024.

\bibitem[Zeng et~al.(2025{\natexlab{a}})Zeng, Huang, Liu, Liu, He, Ma, and He]{zeng2025simplerlzooinvestigatingtamingzero}
Weihao Zeng, Yuzhen Huang, Qian Liu, Wei Liu, Keqing He, Zejun Ma, and Junxian He.
\newblock Simplerl-zoo: Investigating and taming zero reinforcement learning for open base models in the wild, 2025{\natexlab{a}}.
\newblock URL \url{https://arxiv.org/abs/2503.18892}.

\bibitem[Zeng et~al.(2025{\natexlab{b}})Zeng, Huang, Liu, He, Liu, Ma, and He]{zeng2025simplerl}
Weihao Zeng, Yuzhen Huang, Wei Liu, Keqing He, Qian Liu, Zejun Ma, and Junxian He.
\newblock 7b model and 8k examples: Emerging reasoning with reinforcement learning is both effective and efficient.
\newblock \url{https://hkust-nlp.notion.site/simplerl-reason}, 2025{\natexlab{b}}.
\newblock Notion Blog.

\bibitem[Zhong et~al.(2023)Zhong, Cui, Guo, Liang, Lu, Wang, Saied, Chen, and Duan]{zhong2023agieval}
Wanjun Zhong, Ruixiang Cui, Yiduo Guo, Yaobo Liang, Shuai Lu, Yanlin Wang, Amin Saied, Weizhu Chen, and Nan Duan.
\newblock Agieval: A human-centric benchmark for evaluating foundation models, 2023.

\bibitem[Zhou et~al.(2025)Zhou, Wang, Ranjan, Cheng, Tang, He, Liu, and Xing]{zhou2025megamath}
Fan Zhou, Zengzhi Wang, Nikhil Ranjan, Zhoujun Cheng, Liping Tang, Guowei He, Zhengzhong Liu, and Eric~P. Xing.
\newblock Megamath: Pushing the limits of open math corpora.
\newblock \emph{arXiv preprint arXiv:2504.02807}, 2025.
\newblock Preprint.

\end{thebibliography}
\bibliographystyle{colm2024_conference}

\end{document}